\documentclass[11pt]{article}
\usepackage[final]{acl}
\usepackage{times}
\usepackage{latexsym}
\usepackage[T1]{fontenc}
\usepackage[utf8]{inputenc}
\usepackage{microtype}
\usepackage{inconsolata}
\usepackage{graphicx}
\usepackage{tabularx}
\usepackage{enumitem}
\usepackage{makecell}
\usepackage{booktabs}
\usepackage{multirow}
\usepackage[table]{xcolor}
\usepackage{booktabs}
\usepackage{hhline}
\usepackage{adjustbox}
\usepackage[most]{tcolorbox}
\usepackage{amsmath}
\usepackage{natbib}
\newtcolorbox{promptbox}[1]{
  colback=gray!3,
  colframe=black!60,
  colbacktitle=black!65,
  coltitle=white,
  fonttitle=\bfseries,
  fontupper=\small,
  title=#1,
  arc=1mm,
  boxrule=1.2pt,
  left=5pt,
  right=5pt,
  top=3pt,
  bottom=3pt,
  before skip=2pt,
  after skip=2pt,
}
\usepackage{fancyvrb}

\usepackage[most]{tcolorbox}
\tcbuselibrary{listings,breakable}
\usepackage{subcaption}

\usepackage{graphicx}
\usepackage{caption}
\usepackage{booktabs}
\usepackage{multirow}

\usepackage{adjustbox}
\hypersetup{
    colorlinks=true,
    urlcolor=blue
}
\usepackage{xcolor}
\definecolor{stepgreen}{HTML}{61984E}

\title{Don't Count the Edits, Judge by the Outcome Alone:\\Reward-Based Evaluation for Grammatical Error Correction}

\author{
Hayeong Ryu$^{\mathbf{1}}$,
Sunhee Jo$^{\mathbf{2}}$,
Seunguk Yu$^{\mathbf{1}}$,
YoungBin Kim$^{\mathbf{1,2}}$
\\
$^{1}$Department of Artificial Intelligence, Chung-Ang University
\\
$^{2}$Graduate School of Advanced Imaging Sciences, Multimedia and Film, Chung-Ang University
\\
{\ttfamily \{bluebarry37, jo3438, bokju128, ybkim85\}@cau.ac.kr}
}

\begin{document}
\maketitle
\begin{abstract}
Grammatical error correction (GEC) evaluation has traditionally relied on reference or edit overlap, which can penalize valid rewrites that differ from gold corrections. Reference-free metrics reduce this dependence, but evaluating whether a fluent output is a valid correction of the source remains challenging. We propose SURE, a source-conditioned reward evaluator trained on within-source preferences spanning minimal-edit and rewrite-oriented corrections. SURE jointly learns an overall reward with criteria-level supervision for grammaticality, faithfulness, and fluency, together with span-level grounding for source-side error resolution. Experiments on SEEDA show that SURE performs competitively against strong baselines, with particular gains on rewrite-style corrections and more disentangled criteria-level diagnostics.
Our code is available at \url{https://github.com/hayeonggg/SURE}.
\end{abstract}

\section{Introduction}
\label{introduction}

Grammatical error correction (GEC) evaluation has traditionally been defined through reference-based comparison. 
Metrics such as M\textsuperscript{2}~\citep{dahlmeier2012}, ERRANT~\citep{ngetal2014conll, bryant2017ERRANT}, and GLEU~\citep{napolesetal2015ground, tacl_a_00282} operationalize this view by comparing a system output against gold corrections using edit- or surface-level overlap.
This edit- and reference-centered evaluation paradigm, however, introduces several limitations: it can penalize valid corrections that diverge from the gold reference, under-represent the diversity of acceptable corrections, and favor minimal edits over more fluent rewrites~\citep{napoles2017jfleg, chollampatt_2018_reassessment, kobayashi2024SEEDA}. Moreover, edit-wise comparison can miss sentence-level properties such as meaning preservation and naturalness, which are central to whether an output is a valid correction of the source~\citep{choshenabend2018reference}.

\begin{figure}[t!]
    \centering
    \includegraphics[width=\columnwidth, clip, trim=1 2 1 2]{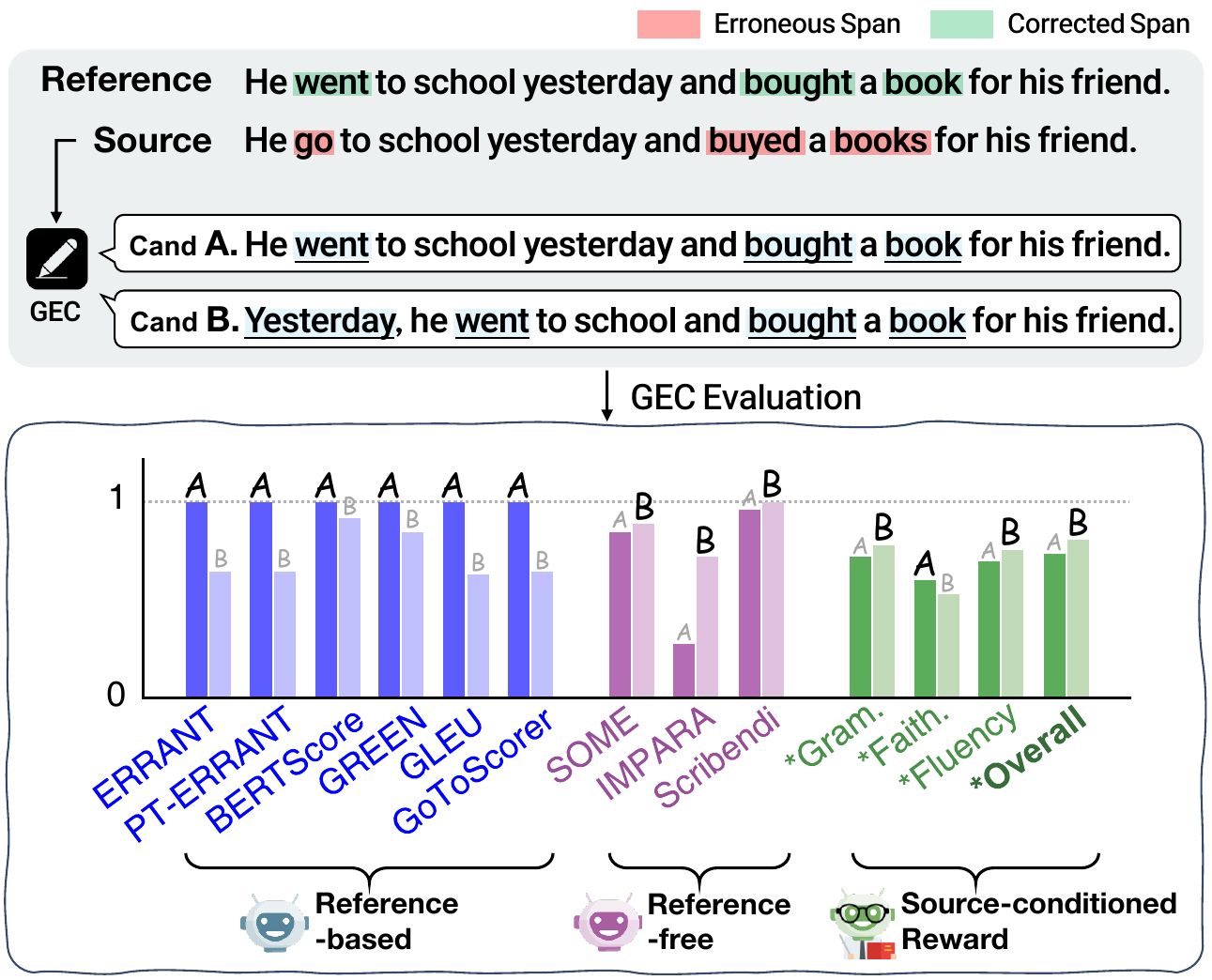}
    \caption{
    Example of evaluation mismatch between reference-based and reference-free metrics for LLM-style correction. 
    The figure compares two candidate corrections for a source sentence and a reference correction, showing how different metric families assign preferences to minimal-edit and rewrite-style outputs. 
    For SURE, Gram., Faith., Flu., and Overall denote grammaticality, faithfulness, fluency, and overall reward, respectively.
    }
    \label{fig:overview}
\end{figure} 

%LLM-style correction에서 reference-based metric과 reference-free metric 사이에 발생하는 evaluation mismatch의 예시. 그림은 source sentence (SR)와 reference correction (RF)에 대해 두 candidate correction을 비교하며, 서로 다른 metric 계열이 local-edit output과 rewrite-style output에 어떤 preference를 부여하는지 보여준다. SURE에서 Gram., Faith., Flu., Overall은 각각 grammaticality, faithfulness, fluency, overall reward를 의미한다.
\begin{figure*}[t!]
    \centering
    \includegraphics[
        width=0.93\textwidth,
        clip,
        trim=3 3 3 3
    ]{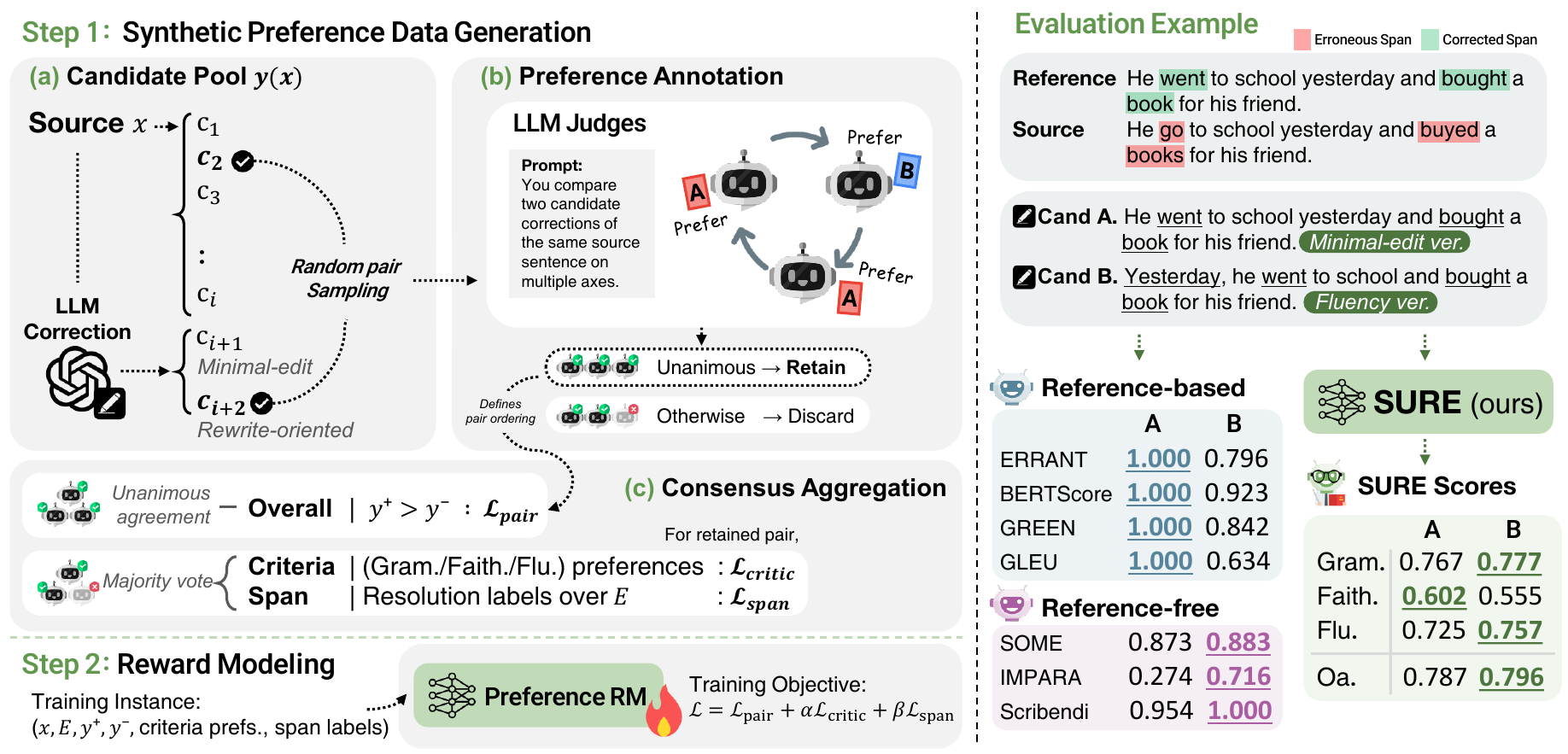}
    \caption{
    Overview of \textbf{SURE}.
    \textbf{(Left)} \textbf{\textcolor{stepgreen}{Step 1:}} Synthetic Preference Data Generation.
    Given a source sentence $x$, we construct a style-diverse candidate pool $\mathcal{Y}(x)$ and aggregate LLM-judge annotations into an overall preference, criteria-level preferences, and span-resolution labels.
    \textbf{\textcolor{stepgreen}{Step 2:}} Reward Modeling.
    We train a source-conditioned reward model with pairwise, criteria-level, and span-level supervision.
    \textbf{(Right)} We compare reference-based and reference-free metrics with SURE using an illustrative evaluation example.
    At inference time, SURE scores candidate corrections using only the source and candidate, without references.
    }
    \label{fig:main}
\end{figure*}

A natural response is to evaluate corrections without relying on a fixed reference. Reference-free and acceptability-based metrics~\citep{asanoetal2017reference, some2020, islammagnani2021end, maedaetal2022impara, kobayashi2024SEEDA} shift the focus from reference overlap to the quality of the source--output relation: whether the output corrects the source, preserves its intended meaning, and reads as grammatical and fluent English.

Existing reference-free metrics already condition on both the source and the correction.
The remaining issue is therefore not source conditioning itself, but how correction validity is supervised.
In particular, prior metrics do not explicitly supervise source-side error resolution or relative preferences among style-diverse corrections of the same source.
Thus, the central challenge is not simply to remove references, but to judge whether a fluent output is justified as a correction of the given source.

% At the same time, removing the reference introduces a different ambiguity. Without an explicit reference, an evaluator must distinguish a good sentence from a good correction: an output may be fluent and natural while omitting source content, changing the intended meaning, or making unnecessary modifications. Thus, the central challenge is not simply to remove references, but to judge whether a fluent output is justified as a correction of the given source. 

This issue is amplified in LLM-based correction, where systems often produce holistic rewrites rather than isolated local edits~\citep{fang2023, katinskaia2024}. Such outputs may improve fluency, naturalness, and style while preserving the source meaning, but they can be penalized by reference-based metrics as reference mismatches or over-corrections~\citep{napoles2017jfleg,kobayashi2024revisiting}. Figure~\ref{fig:overview} illustrates this mismatch. Candidate A closely follows the gold reference and is therefore strongly favored by reference-based metrics, whereas Candidate B rewrites the sentence into a more fluent form and is preferred by several reference-free metrics. Neither behavior is sufficient on its own: evaluation should not reward reference overlap at the expense of valid rewrites, but it should also not treat fluency alone as evidence of correction quality. This motivates training signals that directly capture whether source-side errors are resolved while accounting for grammaticality, faithfulness, and fluency~\citep{some2020,maedaetal2022impara, kobayashi2024SEEDA}.

This discussion leads to a central question: how can GEC evaluation recognize diverse valid corrections while ensuring that the output remains a faithful correction of the source?
We address this question by directly instantiating two requirements of correction validity as training signals: resolving source-side errors and recognizing diverse valid corrections of the same source.
Concretely, we combine span-level error-resolution supervision with preferences between minimal-edit and rewrite-oriented corrections.

We propose SURE, a \textbf{S}ource-conditioned \textbf{U}nified \textbf{R}eward-based \textbf{E}valuator for GEC. 
SURE is trained on within-source preference data spanning minimal-edit and rewrite-oriented corrections, and jointly learns an overall reward with criteria-level supervision and span-level grounding for source-side error resolution.
Experiments on SEEDA~\citep{kobayashi2024SEEDA} show that SURE aligns strongly with human judgments, especially under $+$Fluency and rewrite-style comparisons. Criteria-level analysis further shows that SURE provides more disentangled diagnostic signals than prior multi-criteria evaluators.

Our main contributions are as follows:
\begin{enumerate}[label=(\arabic*), itemsep=0pt, topsep=2pt, parsep=0pt]
    \item We introduce a GEC-specific structured supervision design that combines source-side error-resolution signals with within-source preferences over minimal-edit and rewrite-oriented corrections.
    \item We propose SURE, a reference-free reward evaluator that jointly learns an overall reward with criteria-level and span-level supervision from this preference data.
    \item We show that SURE aligns strongly with human judgments on rewrite-style corrections, and that rewrite-oriented data improves rewrite-related comparisons while introducing a trade-off on traditional GEC pairs.
\end{enumerate}

% This shift raises three basic questions about how GEC evaluation should be framed in the era of LLM-based correction:
% \begin{itemize}
%     \item \textbf{RQ1:} What should count as a valid correction in GEC beyond minimal edits?
%     \item \textbf{RQ2:} How should GEC evaluation account for multiple valid corrections with different surface forms?
%     \item \textbf{RQ3:} What aspects of correction quality should be considered when evaluating rewrite-style corrections?
% \end{itemize}
% To address these questions, we focus on source-conditioned correction validity: whether a candidate correction resolves the source errors, preserves the intended meaning, and produces a natural output.
%§

\section{Problem Formulation}
\label{sec:problem}

\paragraph{Task Formulation.}
Conventional GEC metrics typically score \(y\) by comparing it with a reference correction \(z\), often through edit or surface-form overlap~\citep{dahlmeier2012, napolesetal2015ground, bryant2017ERRANT, tacl_a_00282}. 
For SURE, we adopt a source-conditioned reward formulation, where an evaluator assigns a scalar reward \(R_{\theta}(x,y)\) that reflects the validity of \(y\) as a correction of \(x\). 
This formulation enables reference-free evaluation at inference time, with correction validity as the scoring target.

\paragraph{Correction Validity Criteria.}
Although GEC has traditionally emphasized minimal corrections, the minimal edit is not always the best correction when meaning-preserving rewrites can improve naturalness~\citep{napoles2017jfleg}. 
Prior work has therefore evaluated GEC quality as a multi-dimensional judgment beyond grammatical correctness alone~\citep{asanoetal2017reference, choshenabend2018reference, some2020, maedaetal2022impara}. 
Following this perspective, we define correction validity using three criteria: 
(1) \textbf{Grammaticality} measures whether the candidate is grammatically well-formed and resolves source-side errors without introducing new ones; 
(2) \textbf{Faithfulness} measures whether the candidate preserves the meaning and intent of the source sentence; and 
(3) \textbf{Fluency} measures whether the corrected sentence is natural and idiomatic beyond being merely grammatical. 
These criteria provide the basis for how we characterize correction validity in this work.

\section{SURE: Source-conditioned Unified Reward-based Evaluator} 
\label{Analysis}

We introduce SURE, a reference-free reward evaluator for GEC that scores whether a candidate is a valid correction of the source sentence. 
Figure~\ref{fig:main} provides an overview of the proposed framework, which consists of two stages: synthetic preference data generation and reward modeling. 
We first describe the construction of synthetic preference data from style-diverse correction candidates and consensus among LLM judges.
Then we define a reward model that predicts an overall correction reward and criteria-level scores.

%To construct a benchmark for edit-level validity, we assemble Pair-wise Edit-level Validity Dataset (PEVData) in following three stages.
% We propose SURE, a reference-free GEC evaluation metric based on multi-criteria reward modeling.

\subsection{Synthetic Preference Data Generation} \label{sec:data}
\paragraph{Candidate Pool Construction.}
We first collect source sentences from three standard GEC benchmarks: BEA-2019~\citep{bryantetal2019bea}, CoNLL-2014~\citep{ngetal2014conll}, and JFLEG~\citep{napoles2017jfleg}. We exclude all CoNLL-2014-derived data to avoid source overlap with SEEDA.
To focus on examples that require non-trivial correction judgments, we restrict the source set to sentences with at least 10 tokens and at least two errors identified by ERRANT~\citep{ngetal2014conll, bryant2017ERRANT}.
We further use ERRANT to identify source-side error spans from source--reference pairs, which serve as anchors for span-level annotation.\footnote{
References serve two roles during data construction: they are included as candidate corrections and are used in source--reference pairs to localize source-side errors for span supervision. At inference time, SURE does not require references.
}.
Each candidate pool consists of existing candidates, including human references and GEC system outputs, and two additional corrections generated using GPT-4o~\citep{openai2024gpt4o}.
The GPT-4o-based candidates are generated with two different instructions: one for \textit{minimal-edit correction} and the other for \textit{rewrite-oriented correction}, which encourages a fluent rewrite with a different surface form while preserving the original meaning.
Including rewrite-oriented corrections exposes the preference data to valid corrections that differ substantially from minimal edits.\footnote{The impact of rewrite-oriented corrections is analyzed in Section~\ref{sec:preference}}
The prompts used for LLM-generated corrections are provided in Appendix~\ref{app:prompts}.

\definecolor{oursgreen}{RGB}{232,242,224}

\begin{table*}[t]
\centering
\scriptsize
\setlength{\tabcolsep}{2.3pt}
\renewcommand{\arraystretch}{1.25}

\resizebox{\textwidth}{!}{
\begingroup
\setlength{\arrayrulewidth}{0.10ex}
\setlength{\aboverulesep}{0pt}
\setlength{\belowrulesep}{0pt}
\begin{tabular}{l*{4}{c}!{\vrule width \arrayrulewidth}
                  *{4}{c}!{\vrule width \arrayrulewidth}
                  *{4}{c}!{\vrule width \arrayrulewidth}
                  *{4}{c}}
\toprule

\multirow{4}{*}{\textbf{Metric}}
& \multicolumn{8}{c!{\vrule width \arrayrulewidth}}{\raisebox{-0.25ex}{\textbf{System-level}}}
& \multicolumn{8}{c}{\raisebox{-0.25ex}{\textbf{Sentence-level}}} \\
\hhline{~--------|--------}

& \multicolumn{4}{c!{\vrule width \arrayrulewidth}}{\rule{0pt}{2.3ex}\raisebox{-0.25ex}{\textbf{SEEDA-E}}}
& \multicolumn{4}{c!{\vrule width \arrayrulewidth}}{\rule{0pt}{2.3ex}\raisebox{-0.25ex}{\textbf{SEEDA-S}}}
& \multicolumn{4}{c!{\vrule width \arrayrulewidth}}{\rule{0pt}{2.3ex}\raisebox{-0.25ex}{\textbf{SEEDA-E}}}
& \multicolumn{4}{c}{\rule{0pt}{2.3ex}\raisebox{-0.25ex}{\textbf{SEEDA-S}}} \\

& \multicolumn{2}{c}{\raisebox{-0.20ex}{Base}}
& \multicolumn{2}{c!{\vrule width \arrayrulewidth}}{\raisebox{-0.20ex}{+Fluency}}
& \multicolumn{2}{c}{\raisebox{-0.20ex}{Base}}
& \multicolumn{2}{c!{\vrule width \arrayrulewidth}}{\raisebox{-0.20ex}{+Fluency}}
& \multicolumn{2}{c}{\raisebox{-0.20ex}{Base}}
& \multicolumn{2}{c!{\vrule width \arrayrulewidth}}{\raisebox{-0.20ex}{+Fluency}}
& \multicolumn{2}{c}{\raisebox{-0.20ex}{Base}}
& \multicolumn{2}{c}{\raisebox{-0.20ex}{+Fluency}} \\[-0.15ex]

& ${r}$ & ${\rho}$ & ${r}$ & ${\rho}$
& ${r}$ & ${\rho}$ & ${r}$ & ${\rho}$
& Acc. & $\tau$ & Acc. & ${\tau}$
& Acc. & ${\tau}$ & Acc. & ${\tau}$ \\
\hline

\rowcolor{oursgreen!90}
\multicolumn{17}{l}{\textit{Reference-based}} \\

ERRANT
& 0.386 & 0.364 & -0.576 & -0.130
& 0.242 & 0.084 & -0.656 & -0.305
& 0.629 & 0.257 & 0.568 & 0.135
& 0.588 & 0.177 & 0.517 & 0.034 \\

PT-ERRANT
& 0.588 & 0.545 & -0.536 & 0.024
& 0.464 & 0.273 & -0.617 & -0.147
& 0.613 & 0.226 & 0.554 & 0.109
& 0.591 & 0.181 & 0.517 & 0.034 \\

BERTScore
& -0.031 & 0.552 & -0.513 & -0.002
& -0.046 & 0.294 & -0.549 & -0.165
& 0.563 & 0.125 & 0.500 & -0.001
& 0.558 & 0.115 & 0.477 & -0.046 \\

GREEN
& 0.002 & 0.657 & 0.017 & 0.349
& 0.016 & 0.580 & 0.013 & 0.301
& 0.600 & 0.201 & 0.557 & 0.114
& 0.599 & 0.198 & 0.546 & 0.093 \\

GLEU
& 0.224 & 0.650 & 0.526 & 0.767
& 0.243 & 0.608 & 0.542 & 0.741
& 0.655 & 0.309 & 0.638 & 0.276
& 0.662 & 0.325 & 0.630 & 0.260 \\

GoToScorer
& 0.843 & 0.881 & -0.410 & 0.323
& 0.722 & 0.706 & -0.503 & 0.191
& 0.704 & 0.408 & 0.644 & 0.288
& 0.675 & 0.350 & 0.603 & 0.207 \\

CLEME 2.0
& 0.643 & 0.650 & 0.628 & 0.705
& 0.656 & 0.587 & 0.598 & 0.666
& 0.713 & 0.427 & 0.670 & 0.339
& 0.730 & 0.461 & 0.657 & 0.314 \\

JELV 2.0
& \underline{0.967} & 0.944 & 0.946 & 0.965
& 0.883 & 0.811 & 0.938 & 0.881
& 0.776 & 0.552 & 0.757 & 0.514
& 0.741 & 0.482 & 0.722 & 0.444 \\

\hline
\rowcolor{oursgreen!90}
\multicolumn{17}{l}{\textit{Reference-free}} \\

SOME
& 0.901 & 0.951 & 0.943 & 0.969
& 0.892 & 0.867 & 0.931 & 0.916
& 0.771 & 0.542 & 0.758 & 0.516
& 0.783 & 0.565 & 0.769 & 0.537 \\

Scribendi
& 0.825 & 0.839 & 0.715 & 0.842
& 0.620 & 0.636 & 0.604 & 0.714
& 0.760 & 0.519 & 0.742 & 0.484
& 0.740 & 0.480 & 0.707 & 0.414 \\

IMPARA
& 0.902 & 0.965 & 0.900 & \underline{0.978}
& \underline{0.916} & \textbf{0.902} & 0.887 & \textbf{0.938}
& 0.757 & 0.514 & 0.747 & 0.493
& 0.758 & 0.516 & 0.741 & 0.481 \\

GPT-4.1-E
& 0.584 & 0.602 & 0.893 & 0.746
& 0.520 & 0.634 & 0.874 & 0.766
& 0.741 & 0.482 & 0.742 & 0.483
& 0.733 & 0.465 & 0.747 & 0.494 \\

GPT-4.1-S
& 0.203 & 0.266 & 0.772 & 0.534
& 0.080 & 0.273 & 0.745 & 0.538
& 0.739 & 0.478 & 0.741 & 0.482
& 0.736 & 0.472 & 0.748 & 0.496 \\

GPT-4-S$^{\dagger}$
& 0.960 & 0.958 & 0.967 & 0.969
& 0.887 & 0.860 & 0.931 & 0.908
& \underline{0.798} & \underline{0.595}
& 0.783 & 0.565
& 0.784 & 0.567
& 0.770 & 0.540 \\

\hspace{0.5em}+ Fluency $^{\dagger}$
& \textbf{0.974} & \textbf{0.979}
& \textbf{0.981} & \textbf{0.982}
& 0.913 & 0.874
& \underline{0.952} & 0.916
& \textbf{0.831} & \textbf{0.662}
& \textbf{0.812} & \textbf{0.624}
& \textbf{0.819} & \textbf{0.637}
& \underline{0.797} & \underline{0.594} \\

\hline

\textbf{SURE} (ours)
& 0.932 & \underline{0.972}
& \underline{0.976} & \textbf{0.982}
& \textbf{0.927} & \underline{0.895}
& \textbf{0.970} & \underline{0.934}
& 0.797 & \underline{0.595}
& \underline{0.796} & \underline{0.591}
& \underline{0.809} & \underline{0.619}
& \textbf{0.808} & \textbf{0.615} \\

\bottomrule
\end{tabular}
\endgroup
}

\caption{Meta-evaluation results on SEEDA. Results are reported for both
system-level and sentence-level evaluation under the Base and +Fluency
settings. Metrics are grouped by whether they require references at inference
time. Higher values indicate better agreement with human judgments; the best
score is in bold and the second-best score is underlined.
$^{\dagger}$ indicates scores reported in \cite{kobayashi2024SEEDA}, rather
than results rerun in our experimental setup.}
\label{tab:seeda_results}
\end{table*}

\paragraph{Preference Annotation.}
To construct pairwise preference data, we evaluate correction candidate pairs for each source sentence.
Candidate pairs are randomly sampled from each source's candidate pool and evaluated by three independent LLM judges: GPT-4.1-mini~\citep{openai2025gpt41}, Claude Haiku 4.5~\citep{anthropic2025haiku45}, and Grok 4.3~\citep{xai2026grok43}.
For each comparison, the judges select the better correction overall, assess the candidates along the criteria defined in Section~\ref{sec:problem}, namely grammaticality, faithfulness, and fluency, and annotate whether each marked source error is resolved by each candidate.
We keep only pairs with unanimous agreement on the overall preference, while aggregating criteria-level preferences and span labels by majority vote.
This process yields 2,400 preference pairs from 1,320 unique source sentences.
Filtering statistics are provided in Appendix~\ref{app:data_statistics},
and we validate the filtering strategy with human judgments in
Section~\ref{sec:human_validation}.

%\footnote{\url{https://arxiv.org/abs/2410.21276}}
%\footnote{\url{https://developers.openai.com/api/docs/models/gpt-4.1-mini}}
%\footnote{\url{https://www.anthropic.com/news/claude-haiku-4-5}}
%\footnote{\url{https://docs.x.ai/developers/models/grok-4.3}}

\subsection{Reward Modeling} \label{sec:RM}
\paragraph{Model Architecture.}
We train SURE as a source-conditioned reward model using the preference data constructed in Section~\ref{sec:data}. 
Given a source sentence \(x\) and a candidate correction \(y\), the model encodes the concatenated input \([x; y]\) and predicts an overall reward together with three auxiliary criterion scores $R_{\theta}^{c}(x,y)$ for each $c \in \{\mathrm{gram}, \mathrm{faith}, \mathrm{flu}\}$.
% \begin{equation}
%     R_{\theta}(x,y)=
%     \left(r_{\mathrm{overall}}, r_{\mathrm{gram}}, r_{\mathrm{faith}}, r_{\mathrm{flu}}\right).
% \end{equation}
We use DeBERTa-v3-large~\citep{he2023debertav} as the encoder and fine-tune it with LoRA~\citep{hu2022lora}. 
The overall head is used as the main reward signal, while the criterion heads provide auxiliary supervision for grammaticality, faithfulness, and fluency.

\paragraph{Training Objective.}
For each preference pair \((x, y^{+}, y^{-})\), we optimize the model to score the preferred correction \(y^{+}\) higher than the dispreferred correction \(y^{-}\). 
The training objective combines pairwise reward learning, criteria-level supervision, and span-level grounding:
\begin{equation}
    \mathcal{L}
    =
    \mathcal{L}_{\mathrm{pair}}
    +
    \alpha \mathcal{L}_{\mathrm{critic}}
    +
    \beta \mathcal{L}_{\mathrm{span}} ,
    \label{eq:loss}
\end{equation} 
where $\mathcal{L}_{\mathrm{pair}}$ denotes the pairwise ranking loss over the overall reward, and $\mathcal{L}_{\mathrm{critic}}$ denotes criteria-level pairwise ranking losses for grammaticality, faithfulness, and fluency.
Each criterion-specific loss uses its corresponding preference label, while $\mathcal{L}_{\mathrm{span}}$ is an auxiliary binary classification loss for predicting whether each marked error span in the source is resolved by the candidate correction\footnote{Span annotations are binarized by mapping \textit{resolved} to positive and both \textit{missed} and \textit{partial} to negative, followed by majority voting across judges.}.
The weights \(\alpha\) and \(\beta\) are hyperparameters that balance criteria-level and span-level supervision, respectively.

Span supervision is used only during training: when computing \(\mathcal{L}_{\mathrm{span}}\), we mark error spans in the source so that the encoder learns whether the candidate resolves the corresponding local errors. 
At inference time, the span markers and span head are removed, and SURE takes only the raw source sentence and candidate correction as input.
This allows the model to remain reference-free at evaluation time while using span-level signals to reduce its reliance on surface fluency alone.

\section{Experiments} \label{experiment}

\begin{figure*}[t!]
    \centering

    \begin{subfigure}[t]{0.48\textwidth}
        \centering
        \includegraphics[width=\linewidth, clip, trim=5 5 5 5]{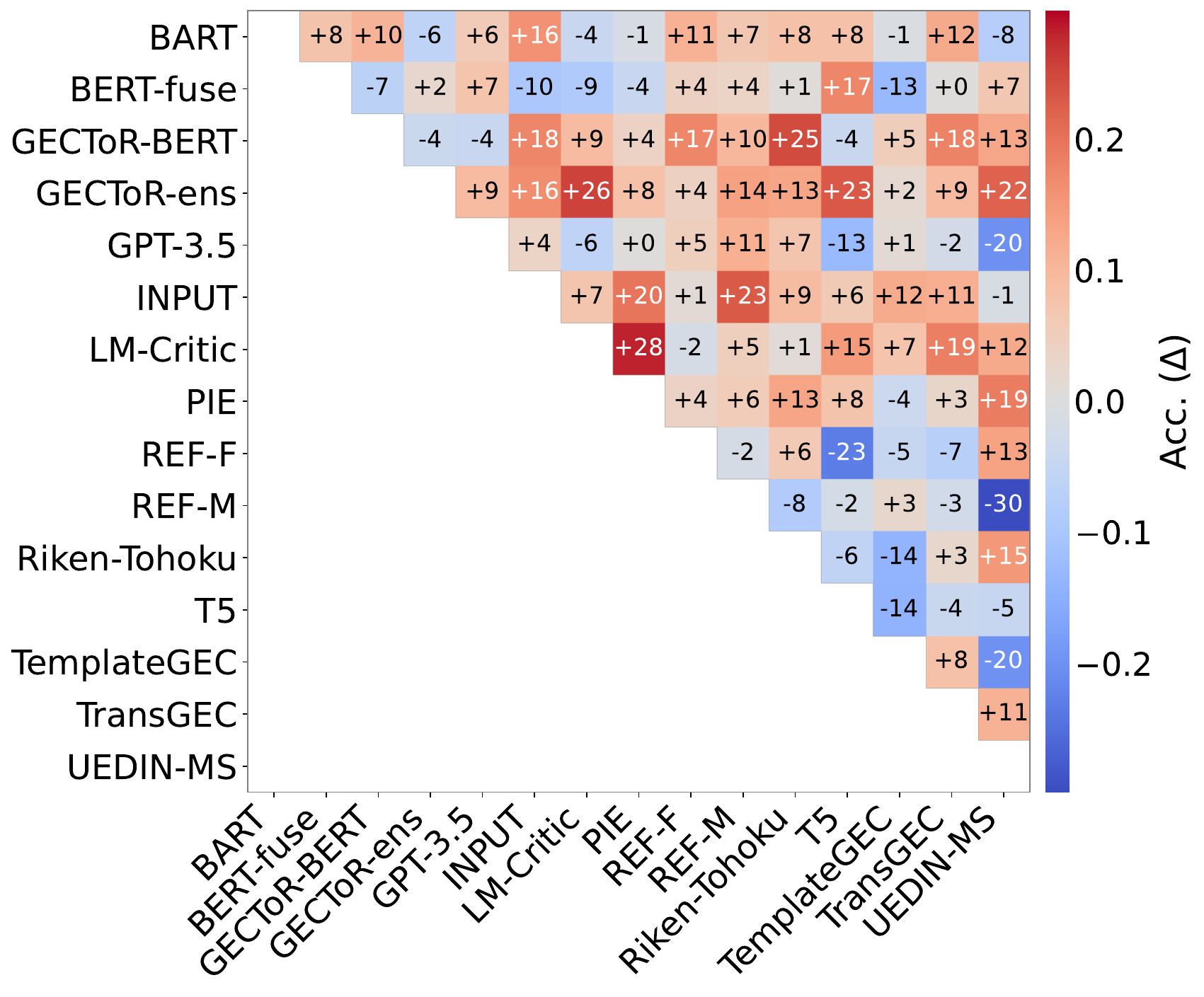}
        \caption{SEEDA-E}
        \label{fig:exp2_heatmap_seeda_e}
    \end{subfigure}
    \hfill
    \begin{subfigure}[t]{0.48\textwidth}
        \centering
        \includegraphics[width=\linewidth, clip, trim=5 5 5 5]{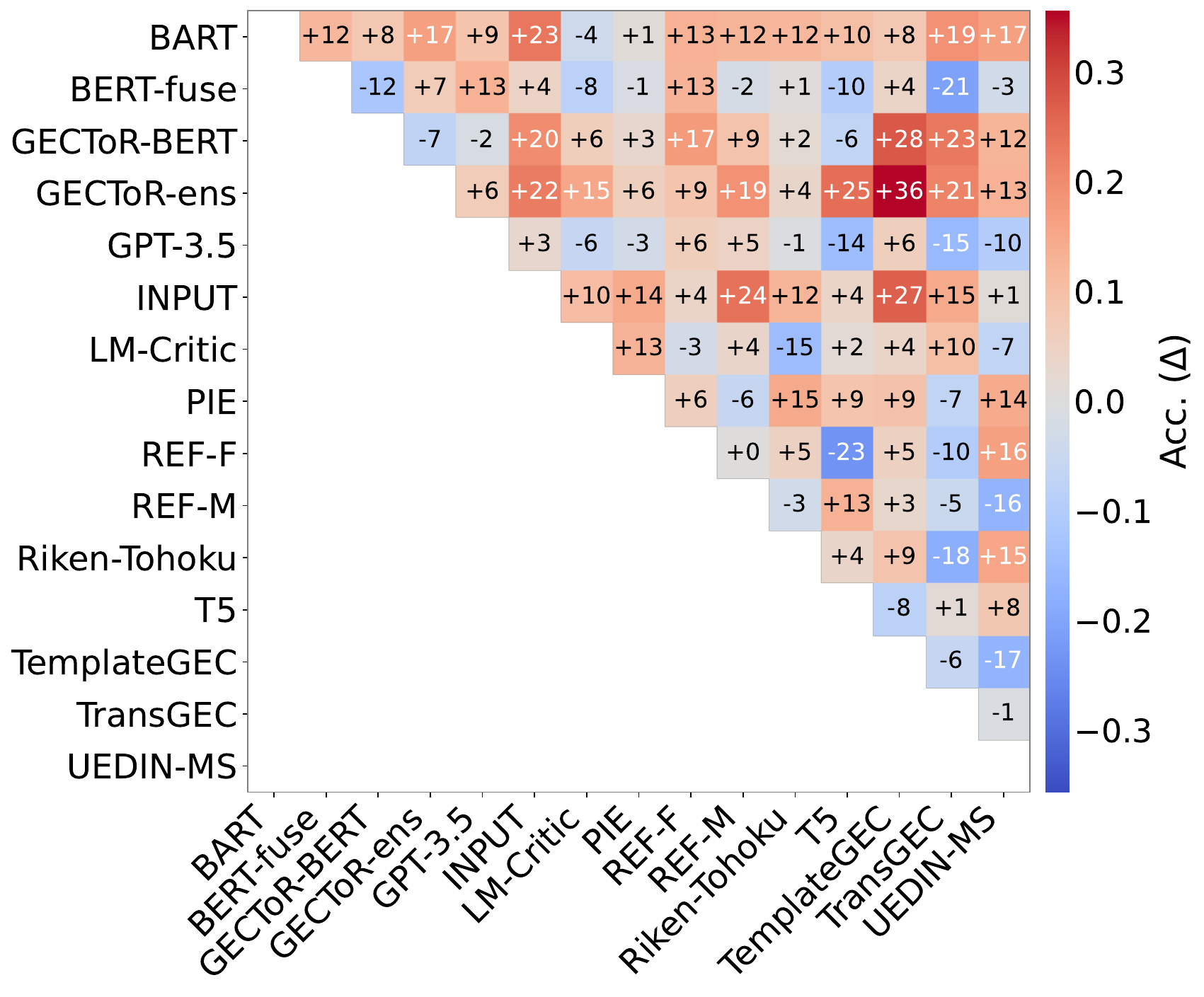}
        \caption{SEEDA-S}
        \label{fig:exp2_heatmap_seeda_s}
    \end{subfigure}

    \caption{
    System-pair accuracy difference between SURE and GPT-4.1-E.
    We select GPT-4.1-E as the comparison system because it is the reference-free LLM evaluator most directly comparable to SURE, assessing edit appropriateness conditioned on both the source and the candidate.
    Each cell reports the pairwise accuracy difference, computed as $\mathrm{Acc.}_{\mathrm{SURE}} - \mathrm{Acc.}_{\mathrm{GPT\text{-}4.1\text{-}E}}$.
    Positive values indicate higher agreement with human preferences by SURE, while negative values indicate higher agreement by GPT-4.1-E.
    }
    \label{fig:pairwise_heatmap}
\end{figure*}

%SEEDA-S에서 SURE와 GPT-4.1-mini-E 간의 system-pair별 accuracy 차이를 보여준다. 각 cell은 pairwise accuracy 차이를 나타내며, 로 계산된다. 양수 값은 SURE가 human preference와 더 잘 일치함을 의미하고, 음수 값은 GPT-4.1-mini-E가 더 잘 일치함을 의미한다. 

\subsection{Experimental Setup} 
\paragraph{Meta-Evaluation.}
We conduct our main evaluation on SEEDA~\citep{kobayashi2024SEEDA}, a meta-evaluation benchmark for GEC metrics.
SEEDA consists of two subsets: SEEDA-E for edit-level pairwise human judgments and SEEDA-S for sentence-level pairwise human judgments. 
Both subsets cover corrections from twelve GEC systems in the Base setting, with two additional fluent human corrections introduced in the $+$Fluency setting.
In this setting, the added fluent corrections test whether a metric can recognize rewrite-style corrections as valid when they preserve meaning and improve naturalness, even if they differ substantially from reference edits.

At the sentence level, we convert metric scores into pairwise preferences and compare them with human judgments using accuracy and Kendall's \(\tau\). 
For system-level evaluation, we aggregate metric scores over corrections for each system and compare the resulting ranking with the human ranking using Pearson's \(r\) and Spearman's \(\rho\).

\begin{table}[t]
\centering
\scriptsize
\setlength{\tabcolsep}{2.2pt}
\renewcommand{\arraystretch}{1.0}

\resizebox{\columnwidth}{!}{%
\begin{tabular}{lcccccccc}
\toprule
\multirow{2}{*}{\textbf{Metric}}
& \multicolumn{4}{c}{\textbf{SEEDA-E}}
& \multicolumn{4}{c}{\textbf{SEEDA-S}} \\
\cmidrule(lr){2-5}
\cmidrule(lr){6-9}

& \textbf{All}
& \textbf{R--T}
& \textbf{R--R}
& \textbf{T--T}
& \textbf{All}
& \textbf{R--T}
& \textbf{R--R}
& \textbf{T--T} \\
\midrule

ERRANT
& 48.2 & 39.8 & 32.4 & 53.8
& 49.8 & 42.0 & 35.8 & 55.7 \\

BERTScore
& 41.0 & 30.4 & 29.6 & 47.4
& 43.8 & 35.9 & 35.5 & 49.4 \\

GLEU
& 51.9 & 52.7 & 40.9 & 51.6
& 55.4 & 57.3 & 41.9 & 54.4 \\
\addlinespace[1pt]
\hline
\addlinespace[2pt]
SOME
& 60.1 & 67.1 & 53.5 & 55.6
& 62.4 & 68.3 & 45.0 & 58.5 \\

IMPARA
& 58.9 & 63.8 & 53.5 & 55.7
& 59.9 & 63.8 & 53.3 & 57.1 \\

GPT-4.1-E
& 60.2 & 71.9 & 53.3 & 54.7
& 59.8 & 69.9 & 52.3 & 54.4 \\
\addlinespace[-1pt]
\midrule
\addlinespace[2pt]
SURE
& \textbf{65.4} & \textbf{73.3} & \textbf{58.1} & \textbf{60.4}
& \textbf{65.7} & \textbf{72.3} & \textbf{58.3} & \textbf{61.0} \\

\bottomrule
\end{tabular}
}
\caption{Pairwise accuracy for rewrite-style correction evaluation. Comparisons are grouped by system style: \textbf{R--T} compares rewrite-style and traditional GEC systems, \textbf{R--R} compares two rewrite-style systems, and \textbf{T--T} compares two traditional GEC systems; \textbf{All} includes all pairs. Values are the percentage of metric preferences matching human preferences, with the best score in each column in \textbf{bold}.}
\label{tab:sub_main}
\end{table}

\paragraph{Baselines.} 
We compare SURE against two groups of GEC evaluation baselines: reference-based metrics and reference-free metrics.
The \textbf{reference-based metrics} include ERRANT~\citep{ngetal2014conll, bryant2017ERRANT}, PT-ERRANT~\citep{gongetal2022}, BERTScore~\citep{Zhang2020BERTScore}, GREEN~\citep{koyamaetal2024}, GLEU~\citep{napolesetal2015ground, tacl_a_00282}, GoToScorer~\citep{gotouetal2020taking}, CLEME 2.0~\citep{yeetal2025cleme2}, and JELV 2.0~\citep{zhan2026jelv}.
The \textbf{reference-free metrics} include SOME~\citep{some2020}, Scribendi~\citep{islammagnani2021end}, IMPARA~\citep{maedaetal2022impara}, and LLM-\{S,E\} evaluators~\citep{kobayashi2024SEEDA}.
For LLM-\{S,E\}, we instantiate the evaluator with GPT-4.1-mini\footnote{https://openai.com/index/gpt-4-1/} and report the edit-level and sentence-level variants as GPT-4.1-E and GPT-4.1-S, respectively.
We additionally include the reported GPT-4-S (+Fluency) results~\citep{kobayashi2024SEEDA}, obtained using GPT-4~\citep{openai2024gpt4technicalreport} (\texttt{gpt-4-1106-preview}).
All non-LLM baselines are evaluated using \texttt{gec-metrics}, a unified evaluation framework for GEC metrics~\citep{gecdemo2025}.

\paragraph{Implementation.}
We implement the model with LoRA using rank 16, scaling factor 32, and dropout 0.1.
The maximum sequence length is set to 256.
We train for 5 epochs with learning rate $2\times10^{-4}$, weight decay 0.01, warmup ratio 0.1, gradient clipping 1.0, and an effective batch size of 8 via gradient accumulation.
The loss weights are set to $\alpha=0.5$ and $\beta=0.2$.

\begin{table}[t]
\centering
\small
\setlength{\tabcolsep}{2.2pt}
\renewcommand{\arraystretch}{1.15}

\begin{adjustbox}{max width=\columnwidth}
\begin{tabular}{llcccc}
\toprule
\multirow{2}{*}{\textbf{Metric}}
& \multirow{2}{*}{\textbf{Setting}}
& \multicolumn{2}{c}{\textbf{SEEDA-E}}
& \multicolumn{2}{c}{\textbf{SEEDA-S}} \\
\cmidrule(lr){3-4}
\cmidrule(lr){5-6}
& & Base & +Fluency
& Base & +Fluency \\
\midrule

\multirow{2}{*}{\makecell{System\\--level\\($r$)}}
& \makecell{w/o rewrite\\[0.3ex]}
& 0.910 & 0.967 & 0.922 & 0.962 \\
& \makecell{w/\hspace{0.7em}rewrite\\[0.2ex]\phantom{(+0.000)}}
& \makecell{0.926\\\textcolor{red}{(+0.016)}}
& \makecell{0.978\\\textcolor{red}{(+0.011)}}
& \makecell{0.945\\\textcolor{red}{(+0.023)}}
& \makecell{0.989\\\textcolor{red}{(+0.027)}} \\

\midrule

\multirow{2}{*}{\makecell{Sentence\\--level\\(Acc.)}}
& \makecell{w/o rewrite\\[0.3ex]}
& 0.808 & 0.802 & 0.820 & 0.821 \\
& \makecell{w/\hspace{0.7em}rewrite\\[0.2ex]\phantom{(+0.000)}}
& \makecell{0.791\\\textcolor{blue}{(-0.017)}}
& \makecell{0.785\\\textcolor{blue}{(-0.017)}}
& \makecell{0.822\\\textcolor{red}{(+0.002)}}
& \makecell{0.821\\\textcolor{gray}{(+0.000)}} \\

\bottomrule
\end{tabular}
\end{adjustbox}

\caption{Effect of rewrite-oriented preference data. w/o rewrite denotes training without rewrite candidates, while w/ rewrite denotes training with rewrite candidates. Values in parentheses indicate the difference from w/o rewrite.
Ablation models are retrained under a controlled setting, so their absolute scores are not intended to exactly match the final SURE scores in Table~\ref{tab:seeda_results}; we focus on relative differences between ablation conditions.}
\label{tab:rewrite_effect}
\end{table}

%Rewrite-oriented preference data의 효과를 분석한 결과이다. \textit{w/o rewrite}는 rewrite candidate 없이 학습한 설정을, \textit{w/ rewrite}는 rewrite candidate를 포함해 학습한 설정을 의미한다. 괄호 안 값은 \textit{w/o rewrite} 대비 차이를 나타낸다.

\subsection{Correction Validity Evaluation} \label{sec:main_result}
We evaluate each metric against human judgments on SEEDA. As shown in Table~\ref{tab:seeda_results}, reference-based metrics perform reasonably in the Base setting, but their agreement often drops under $+$Fluency, where valid corrections may diverge from reference edits. Reference-free metrics are generally more stable in this setting, indicating the advantage of evaluating correction validity without relying on reference overlap.
Among reference-free metrics, SURE maintains strong agreement across both Base and $+$Fluency settings at the system and sentence levels. The results indicate that source-conditioned reward estimation provides a robust criterion for evaluating corrections beyond reference overlap.
Additional results on cross-domain transfer are provided in Appendix~\ref{app:gmeg_transfer}.

\begin{figure}[t!]
    \centering
    \includegraphics[width=\columnwidth, clip, trim=5 5 5 5]{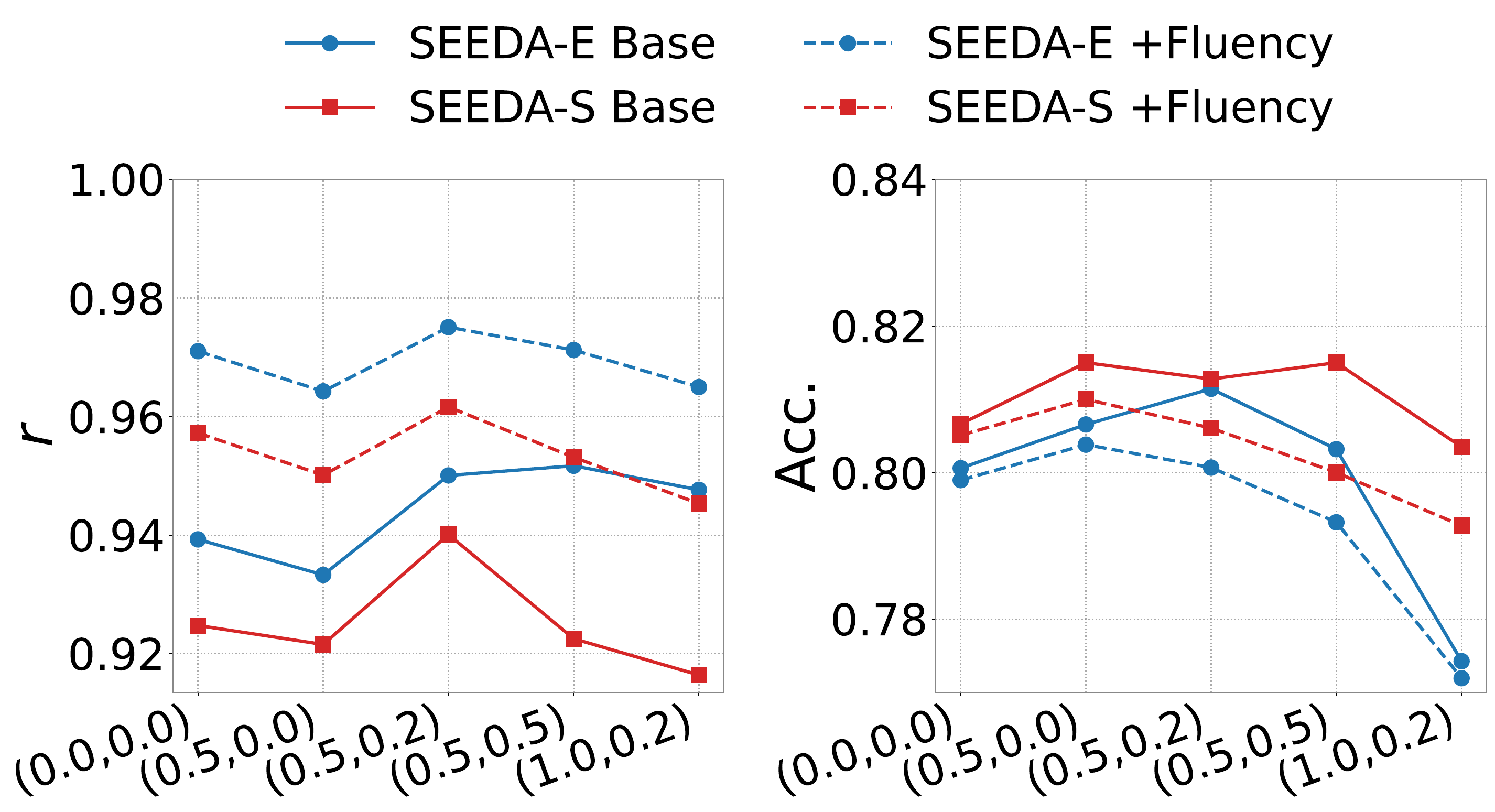}
    \caption{Sensitivity to auxiliary loss weights $\alpha$ and $\beta$ in Eq.~\ref{eq:loss}. We report system-level Pearson's $r$ and sentence-level accuracy (Acc.) on SEEDA-E and SEEDA-S under Base and $+$Fluency settings. The default setting is $(\alpha,\beta)=(0.5,0.2)$.}
    \label{fig:hyper}
\end{figure} 

\begin{table}[t]
\centering
\small
\setlength{\tabcolsep}{3pt}
\renewcommand{\arraystretch}{1.08}

\begin{adjustbox}{max width=\columnwidth}
\begin{tabular}{llcl}
\toprule
\textbf{Dataset} & \textbf{Pair Type} & \textbf{w/o rewrite} & \textbf{w/ rewrite} \\
\midrule

\multirow{4}{*}{SEEDA-E}
& All
& 67.30 & 66.68 \textcolor{blue}{(-0.62pp)} \\
& Trad. only
& 61.03 & 58.77 \textcolor{blue}{(-2.26pp)} \\
& Rewrite-inv.
& 76.50 & 78.30 \textcolor{red}{(+1.80pp)} \\
& Rewrite vs. Trad.
& 77.15 & 79.20 \textcolor{red}{(+2.05pp)} \\

\midrule

\multirow{4}{*}{SEEDA-S}
& All
& 66.52 & 65.45 \textcolor{blue}{(-1.07pp)} \\
& Trad. only
& 61.07 & 57.39 \textcolor{blue}{(-3.68pp)} \\
& Rewrite-inv.
& 73.41 & 75.64 \textcolor{red}{(+2.23pp)} \\
& Rewrite vs. Trad.
& 74.26 & 76.37 \textcolor{red}{(+2.11pp)} \\

\bottomrule
\end{tabular}
\end{adjustbox}

\caption{Pair-type analysis of rewrite-oriented preference data. w/o rewrite denotes training without rewrite candidates, while w/ rewrite denotes training with rewrite candidates. Trad. only compares two traditional GEC systems, Rewrite-inv. includes pairs where at least one system is rewrite-style, and Rewrite vs. Trad. compares a rewrite-style system with a traditional GEC system. Values are pairwise accuracy (\%) against human preferences.}
\label{tab:rewrite_pair_ablation}
\end{table}

\section{Analysis} \label{sec:analysis} 
\subsection{Evaluation Behavior of SURE} \label{sec:Ablation}
\paragraph{Rewrite-Style Correction Evaluation.}
We further group SEEDA pairs by comparison type and evaluate which metric best matches human preferences within each group, as shown in Table~\ref{tab:sub_main}. SURE achieves the highest accuracy across all groups on both SEEDA-E and SEEDA-S. The separation is most visible in rewrite-related comparisons, where edit-overlap metrics fall behind reference-free metrics. This result shows that SURE is better aligned with human judgments when evaluating rewrite-style corrections, while remaining competitive on traditional GEC comparisons. %We provide a system-level pairwise comparison with GPT-4.1-E in Appendix~\ref{app:pairwise_heatmap}, showing that the observed advantage is not driven by a small subset of comparisons.

\paragraph{System-Pair Analysis.}
We further compare SURE with GPT-4.1-E at the system-pair level, as shown in Figure~\ref{fig:pairwise_heatmap}. Positive differences appear across many system pairs, indicating that SURE's advantage is not limited to aggregate scores. The gains are especially frequent around rewrite-style systems such as GPT-3.5 and REF-F, while negative cells remain localized to a small number of specific pairs.

\paragraph{Auxiliary supervision.}
We analyze whether the auxiliary criteria-level and span-level losses contribute to the overall reward model, varying their loss weights as summarized in Figure~\ref{fig:hyper}. The setting $(0.0,0.0)$ corresponds to the pairwise-only model, while $(0.5,0.0)$ removes span-level supervision and isolates the effect of criteria-level learning. The results show that SURE is not highly sensitive to the exact choice of these weights, with stable performance across both system- and sentence-level evaluations. At the same time, using auxiliary supervision generally improves or preserves performance compared to the pairwise-only objective, indicating that criteria- and span-level signals provide useful grounding for the overall reward. 

We further examine the system-level consequences of these evaluation mismatches under the $+$Fluency setting; complete ranking analyses are provided in Appendix~\ref{sec:system_ranking}.

% \begin{table*}[t]
% \centering
% \scriptsize
% \setlength{\tabcolsep}{3.5pt}
% \renewcommand{\arraystretch}{1.05}
% \resizebox{0.85\textwidth}{!}{
% \begin{tabular}{l*{8}{c}}
% \toprule
% \multirow{2}{*}{\textbf{Judge/Aggregator}}
% & \multicolumn{2}{c}{\textbf{Overall}}
% & \multicolumn{2}{c}{\textbf{Grammaticality}}
% & \multicolumn{2}{c}{\textbf{Faithfulness}}
% & \multicolumn{2}{c}{\textbf{Fluency}} \\
% \cline{2-9}
% & Acc. & $\kappa$
% & Acc. & $\kappa$
% & Acc. & $\kappa$
% & Acc. & $\kappa$ \\
% \hline

% GPT-4.1-mini
% & 0.760 & 0.515
% & 0.770 & 0.536
% & 0.395 & -0.200
% & 0.770 & 0.533 \\

% Claude-Haiku-4.5
% & 0.805 & 0.606
% & 0.765 & 0.527
% & \textbf{0.685} & \textbf{0.366}
% & 0.820 & 0.635 \\

% Grok-4.3
% & 0.755 & 0.509
% & 0.720 & 0.440
% & 0.470 & -0.052
% & 0.805 & 0.607 \\

% Majority 
% & 0.790 & 0.577
% & 0.745 & 0.487
% & 0.490 & -0.018
% & 0.810 & 0.614 \\

% Unanimous 
% & \textbf{0.831} & \textbf{0.658}
% & \textbf{0.819} & \textbf{0.635}
% & 0.561 & 0.129
% & \textbf{0.842} & \textbf{0.677} \\

% \bottomrule
% \end{tabular}
% }
% \caption{Agreement with human annotations across judge models and aggregation strategies.}
% \label{tab:judge_aggregator_agreement}
% \end{table*}

\begin{table}[t]
\centering
\small
\setlength{\tabcolsep}{1.8pt}
\renewcommand{\arraystretch}{1.05}

\begin{adjustbox}{max width=\columnwidth}
\begin{tabular}{l*{8}{c}}
\toprule
\multirow{2}{*}{\makecell{\textbf{Judge}/\\\textbf{Aggregator}}}
& \multicolumn{2}{c}{\textbf{Overall}}
& \multicolumn{2}{c}{\textbf{Gram.}}
& \multicolumn{2}{c}{\textbf{Faith.}}
& \multicolumn{2}{c}{\textbf{Flu.}} \\
\cline{2-9}
\addlinespace[1pt]
& Acc. & $\kappa$
& Acc. & $\kappa$
& Acc. & $\kappa$
& Acc. & $\kappa$ \\
\addlinespace[2pt]
\hline
\addlinespace[3pt]
GPT-4.1-mini
& 0.760 & 0.515
& 0.770 & 0.536
& 0.395 & -0.200
& 0.770 & 0.533 \\

Claude Haiku 4.5
& 0.805 & 0.606
& 0.765 & 0.527
& \textbf{0.685} & \textbf{0.366}
& 0.820 & 0.635 \\

Grok 4.3
& 0.755 & 0.509
& 0.720 & 0.440
& 0.470 & -0.052
& 0.805 & 0.607 \\
\addlinespace[1pt]
\hline
\addlinespace[1pt]
Majority
& 0.790 & 0.577
& 0.745 & 0.487
& 0.490 & -0.018
& 0.810 & 0.614 \\

Unanimous
& \textbf{0.831} & \textbf{0.658}
& \textbf{0.819} & \textbf{0.635}
& 0.561 & 0.129
& \textbf{0.842} & \textbf{0.677} \\

\bottomrule
\end{tabular}
\end{adjustbox}

\caption{Reliability of LLM-based preference annotations on SEEDA pairs. The first three rows report individual judge performance. Majority aggregates the three LLM judges by majority vote, while unanimous retains only pairs on which all judges agree.}
\label{tab:judge_reliability}
\end{table}

%200개의 SEEDA pair에서 LLM 기반 preference annotation의 신뢰도를 평가한 결과이다. 상위 세 행은 개별 judge의 성능을 나타낸다. Majority는 세 LLM judge의 다수결 결과이고, unanimous는 세 judge가 모두 일치한 pair만 사용한 결과이다.

\subsection{Effect and Reliability of Preference Data} \label{sec:preference}
\paragraph{Effect of Rewrite-Oriented Data.}
We ablate rewrite-oriented candidates during preference-data construction to examine their contribution to SURE (Table~\ref{tab:rewrite_effect}). Adding rewrite data consistently improves system-level correlation across both SEEDA-E and SEEDA-S, with larger gains in the $+$Fluency setting. This indicates that style-diverse preference data helps the model rank systems more accurately when fluent rewrite-style corrections are included. Sentence-level accuracy is largely preserved, although SEEDA-E shows a small decrease, suggesting a mild trade-off with edit-level judgments that favor more local corrections.

\paragraph{Pair-Type Effects of Rewrite-Oriented Data.}
Table~\ref{tab:rewrite_pair_ablation} further breaks down sentence-level accuracy by system style. 
Comparing models trained with and without rewrite candidates, we find that rewrite-oriented data improves pairs involving rewrite-style systems, especially Rewrite vs. Trad. comparisons, but reduces accuracy on Trad. only pairs. 
This explains the small change in aggregate sentence-level accuracy: gains on rewrite-related comparisons are partially offset by drops on traditional-only comparisons. 
Thus, rewrite-oriented data mainly improves the evaluation of rewrite-style systems, while introducing a mild trade-off for minimal-edit-oriented comparisons.

\paragraph{Reliability of Consensus-Based Preference Data.}
The quality of preference data depends on whether LLM judges provide judgments that are consistent with human preferences. Table~\ref{tab:judge_reliability} shows that enforcing unanimous agreement improves reliability over both individual judges and majority voting, yielding the highest overall accuracy and $\kappa$. This provides empirical support for using unanimous consensus as the filtering criterion for constructing training pairs.

\begin{figure}[t!]
    \centering
    \includegraphics[width=\columnwidth, clip, trim=5 5 5 5]{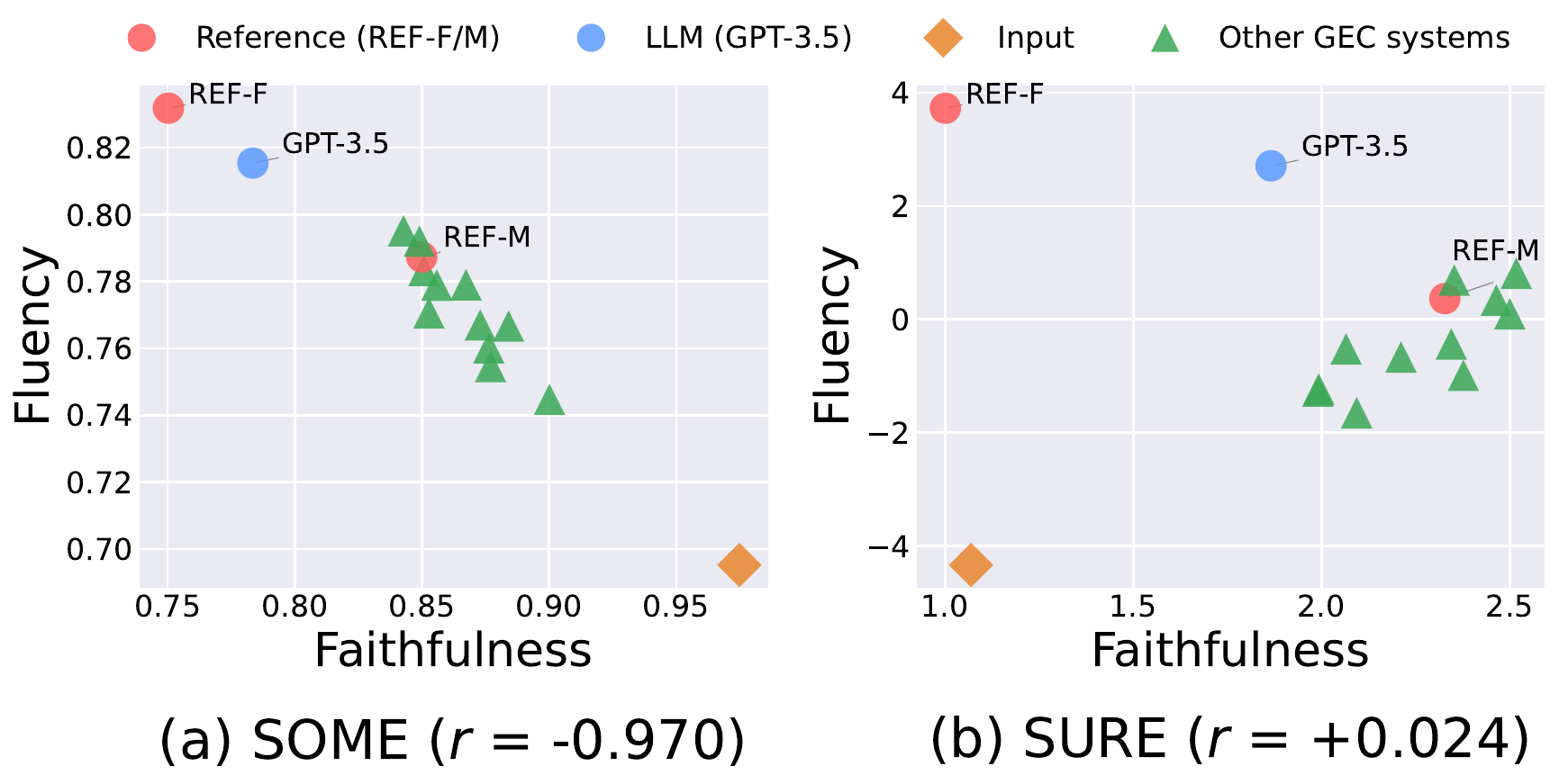}
    \caption{Criteria-level diagnostics of faithfulness and fluency. Each point represents a GEC system, with scores averaged over its outputs. For SOME, the meaning-preservation score is treated as faithfulness for consistency. Pearson's $r$ measures the correlation between the two criteria.}
    \label{fig:criteria_scatter}
\end{figure} 

%Faithfulness와 fluency에 대한 criteria-level diagnostics이다. 각 점은 하나의 GEC system을 나타내며, score는 해당 system의 output에 대해 평균낸 값이다. SOME의 경우, 일관성을 위해 meaning-preservation score를 faithfulness로 간주한다. Pearson’s r은 두 criteria 간 상관을 측정한다.

\begin{table}[t]
    \centering
    \small
    \setlength{\tabcolsep}{3pt}
    \begin{tabularx}{\columnwidth}{@{}Xcc@{}}
    \toprule
    \raisebox{3pt}{\textbf{Metric}} & \textbf{\shortstack{Retained\\(3--0)}}
           & \textbf{\shortstack{Discarded\\(2--1)}} \\
    \midrule
    A/B consensus
        & 62.0\% (31/50) & 38.0\% (19/50) \\
    Human pairwise agreement
        & 49.7\% & 41.7\% \\
    LLM--Human (Overall)
        & 77.4\% (24/31) & 57.9\% (11/19) \\
    \bottomrule
    \end{tabularx}
    \caption{
    Human validation of retained and discarded preference pairs.
    A/B consensus denotes the proportion of pairs for which human annotators
    reached consensus, and Human pairwise agreement is the raw percent agreement across all six annotator pairs on the three-way Overall judgment (A, B, or Difficult).
    LLM--Human (Overall) measures agreement between the LLM preference
    and the human consensus judgment.
    Retained pairs received unanimous 3--0 LLM judgments, whereas
    discarded pairs received 2--1 split judgments.
    }
    \label{tab:human_validation}
\end{table}
\begin{table}[t]
    \centering
    \small
    \begin{tabular}{lcc}
    \toprule
    \textbf{Criterion} & \textbf{Agreement} & \textbf{Cases} \\
    \midrule
    Overall & 77.4\% & 24/31 \\
    Gram.   & 92.3\% & 12/13 \\
    Faith.  & 84.0\% & 21/25 \\
    Flu.    & 75.9\% & 22/29 \\
    \bottomrule
    \end{tabular}
    \caption{
    LLM--Human agreement on retained preference pairs.
    Agreement is computed over pairs with human consensus for each dimension.
    Cases report the number of LLM--Human agreements over the number of
    human-consensus pairs.
    }
    \label{tab:human_criteria}
\end{table}
\begin{table}[t]
    \centering
    \small
    \begin{tabularx}{\columnwidth}{Xr}
    \toprule
    \textbf{Measure} & \textbf{Value} \\
    \midrule
    Human rewrite preference
        & 83.3\% \\
    LLM rewrite preference
        & 66.7\% \\
    Human agreement with LLM rewrite
        & 87.5\% (7/8) \\
    \bottomrule
    \end{tabularx}
    \caption{
    Rewrite preference analysis on human-consensus rewrite--non-rewrite pairs.
    The first two rows report the rewrite preference rates of human annotators
    and LLM judges, respectively.
    The last row reports human agreement among cases in which the LLM judges
    preferred the rewrite.
    }
    \label{tab:rewrite_bias}
\end{table}

\subsection{Criteria-Level Diagnostics}
We evaluate whether the criteria-level scores offer diagnostic information beyond the overall reward. A useful multi-criteria evaluator should distinguish different correction profiles, such as fluency-oriented rewrites and meaning-preserving conservative corrections. As shown in Figure~\ref{fig:criteria_scatter}, SOME~\cite{some2020} exhibits a strong negative correlation between faithfulness and fluency, indicating that the two scores largely behave as opposite ends of a single axis. \textcolor{black}{By contrast, SURE yields nearly independent faithfulness and fluency scores, allowing these correction profiles to be separated in the criteria space. A similar separation is observed between grammaticality and faithfulness, while grammaticality and fluency remain closely related (see Appendix~\ref{app:criteria_scatter}). In Appendix~\ref{app:tmu_gfm}, we further evaluate criteria-level transfer on TMU-GFM; SURE shows reasonable zero-shot transfer on confident grammaticality and fluency pairs, while meaning preservation remains challenging.}

\subsection{Human Validation of Preference Data}
\label{sec:human_validation}

\paragraph{Human Evaluation Setup.}
To assess whether the LLM-generated preference labels align with independent
human judgments, we conduct a human evaluation with four annotators on
100 candidate pairs: 50 retained pairs with 3--0 unanimous LLM agreement
and 50 discarded pairs with 2--1 split judgments.
The pairs are approximately evenly distributed across the three pair types.
Human consensus is defined as a strict majority of at least three of four
valid votes.

\paragraph{Effect of Unanimous Filtering.}
As shown in Table~\ref{tab:human_validation}, retained pairs exhibit higher human consensus, pairwise agreement, and LLM--Human agreement than discarded pairs.
In particular, the human consensus rate increases from 38.0\% for discarded
pairs to 62.0\% for retained pairs, and the difference is statistically
significant ($p=0.027$).
Among pairs for which humans reach consensus, the retained LLM labels agree
with the human overall preference in 77.4\% of cases, compared with 57.9\%
for discarded pairs.
These results suggest that unanimous filtering preferentially retains
less ambiguous comparisons and improves the human alignment of the
preference labels used for training.

% 학습에 사용된 retained llm labels가 human judgment와 criterion별로 얼마나 일치하는가?
\paragraph{Criteria-Level Agreement.}
We further compare the retained LLM labels with human consensus separately
for each evaluation criterion.
Table~\ref{tab:human_criteria} shows that agreement is highest for
grammaticality (92.3\%), followed by faithfulness (84.0\%) and fluency
(75.9\%).
Together with the 77.4\% overall agreement, these results indicate that
the supervision used to train SURE is broadly consistent with independent
human judgments across both the overall and criteria-level labels.

\paragraph{Audit of Rewrite Preference Bias.}
We further assess whether LLM-judge supervision introduces a systematic preference for rewrite-style corrections.
As shown in Table~\ref{tab:rewrite_bias}, among rewrite--non-rewrite pairs with human consensus, the rewrite preference rate is higher for human annotators than for the LLM judges (83.3\% vs.\ 66.7\%).
Furthermore, human judgments agree with seven of the eight cases in which the LLM judges prefer the rewrite.
These results suggest that the observed rewrite preferences are not driven by a systematic over-preference for rewrites by the LLM judges.

\section{Related Work}
\label{related_work}
\subsection{Reference-Based Metrics for GEC}
Reference-based metrics have long served as the dominant framework for GEC evaluation. 
M\textsuperscript{2} and ERRANT compare extracted system edits against gold edits~\citep{dahlmeier2012,bryant2017ERRANT}. 
GLEU evaluates corrections through n-gram overlap with references and the source sentence~\citep{napolesetal2015ground}. 
Subsequent metrics refine this paradigm through contextual or pretrained scoring, as in BERTScore and PT-ERRANT~\citep{Zhang2020BERTScore,gongetal2022}; alignment-free comparison, as in GREEN~\citep{koyamaetal2024}; or difficulty-aware edit weighting, as in GoToScorer~\citep{gotouetal2020taking}.

Recent work has further improved interpretability and validity within the edit-based framework. 
CLEME 2.0 decomposes edit outcomes into hit-, wrong-, under-, and over-correction~\citep{yeetal2025cleme2}, while JELV introduces an edit-level validity judge for evaluation and reference expansion~\citep{zhan2026jelv}. 
Meta-evaluation studies have also shown that conclusions from reference-based metrics can depend on system sets, human-evaluation granularity, and reference coverage~\citep{chollampatt_2018_reassessment,kobayashi2024SEEDA}. 
These studies make reference-based evaluation more robust and interpretable, but the target remains tied to gold corrections or reference-derived edits. 
We evaluate whether a candidate is a valid correction of the source without requiring reference overlap at inference time.

\subsection{Reference-Free and LLM-Based GEC Evaluation}

Reference-free metrics evaluate GEC outputs without comparing them against gold corrections, reducing dependence on reference coverage. 
SOME decomposes reference-free evaluation into grammaticality, fluency, and meaning preservation~\citep{some2020}, while Scribendi proposes a straightforward reference-free alternative to gold-standard comparison~\citep{islammagnani2021end}. 
IMPARA estimates correction quality from the impact of a candidate correction on the source sentence using parallel grammatical and ungrammatical sentence pairs~\citep{maedaetal2022impara}. 

These approaches move beyond surface overlap and already condition evaluation on the source and candidate correction.
However, they do not explicitly combine source-side error-resolution supervision with relative preferences among style-diverse corrections of the same source.
Recent work has also explored LLMs as reference-free evaluators for GEC~\citep{kobayashi2024SEEDA}, using edit-level and sentence-level prompts and showing that LLMs can capture broader aspects of correction quality than overlap-based metrics.
Our work instead focuses on this supervision design, combining source-side error-resolution signals with preferences among minimal-edit and rewrite-oriented corrections.

\subsection{Preference and Reward Modeling for Evaluation}

Preference-based reward modeling learns evaluators from pairwise judgments by assigning higher scores to preferred outputs, a formulation widely used in language generation and alignment~\citep{Christiano2017,ziegler2020,Stiennon2020,ouyang2022training,bai2022traininghelpfulharmlessassistant,rafailov2023direct}. 
Learned text generation metrics such as BLEURT, COMET, BARTScore, and UniEval further capture semantic and quality-oriented judgments beyond surface overlap~\citep{sellametal2020bleurt,rei2020comet,NEURIPS2021_e4d2b6e6,zhong2022towards}.

Recent work uses LLMs as judges to collect preferences or directly evaluate generated outputs~\citep{liu2023geval,wang2023chatgpt,zheng2023judging,chen2024alpagasus, fu2023gptscore, dubois2024lengthcontrolled}. 
While such approaches provide scalable supervision, their judgments can vary with prompts, judge models, and evaluation criteria~\citep{zheng2023judging,liu2023geval,wang2024rewardbench}. 
For correction evaluation, preferences should reflect whether an output is justified as a correction of the given source, rather than whether it is merely fluent or well-written. 
We adopt preference-based evaluation in this source-conditioned setting, assessing correction quality through grammaticality, faithfulness, fluency, and span-level error resolution.

% Preference-based reward modeling learns evaluators from pairwise judgments by assigning higher scores to preferred outputs than to dispreferred ones. 
% This formulation has been widely used to learn reward functions from human feedback in language generation and alignment~\citep{Christiano2017,ziegler2020,Stiennon2020,ouyang2022training,bai2022traininghelpfulharmlessassistant,rafailov2023direct}. 
% In text generation evaluation, learned metrics such as BLEURT, COMET, BARTScore, and UniEval have also shown that semantic and quality-oriented judgments can be captured beyond surface overlap~\citep{sellametal2020bleurt,rei2020comet,NEURIPS2021_e4d2b6e6,zhong2022towards}.

% Recent work uses LLMs as judges to collect preferences or directly evaluate generated outputs~\citep{fu2023gptscore,liu2023geval,wang2023chatgpt,zheng2023judging,chen2024alpagasus,dubois2024lengthcontrolled}. 
% While such approaches provide scalable supervision, their judgments can vary with prompts, judge models, and evaluation criteria~\citep{zheng2023judging,liu2023geval,wang2024rewardbench}. 
% For correction evaluation, preferences should reflect whether an output is justified as a correction of the given source, rather than whether it is merely fluent or well-written. 
% We adopt preference-based evaluation in this source-conditioned setting, assessing correction quality through grammaticality, faithfulness, fluency, and span-level error resolution.

\section{Conclusion}
\label{conclusion}
We presented SURE, a source-conditioned reward evaluator for grammatical error correction.
SURE learns correction validity from style-diverse preferences with criteria-level and span-level supervision.
Experiments on SEEDA show strong alignment with human judgments, particularly for fluent rewrite-style corrections, while providing useful diagnostics for grammaticality, faithfulness, and fluency.

\section*{Limitations}
This work has several limitations. First, SURE is trained on synthetic and LLM-annotated preference data, and may inherit biases from the judge models used during data construction. 
Second, SURE exhibits uneven cross-domain generalization, with weaker system-level correlations on Wiki despite competitive transfer on FCE and at the sentence level. This suggests that its calibration may remain sensitive to domain shift, motivating broader validation across domains and languages.
Third, although references are not required at inference time, reference-derived ERRANT spans are used during training for span-level supervision. Future work will expand human-annotated preference data, evaluate multilingual settings, and improve the calibration of criteria-level rewards.

\section*{Acknowledgments}
This work was supported by the Institute of Information \& Communications Technology Planning \& Evaluation (IITP) grant funded by the Korea government (MSIT) [RS-2021-II211341, Artificial Intelligence Graduate School Program (Chung-Ang University)] and by the National Research Foundation of Korea (NRF) grant funded by the Korea government (MSIT) (RS-2025-00556246).

\bibliography{custom}

@inproceedings{bryantetal2019bea,
    title = "The {BEA}-2019 Shared Task on Grammatical Error Correction",
    author = "Bryant, Christopher  and
      Felice, Mariano  and
      Andersen, {\O}istein E.  and
      Briscoe, Ted",
    editor = "Yannakoudakis, Helen  and
      Kochmar, Ekaterina  and
      Leacock, Claudia  and
      Madnani, Nitin  and
      Pil{\'a}n, Ildik{\'o}  and
      Zesch, Torsten",
    booktitle = "Proceedings of the Fourteenth Workshop on Innovative Use of NLP for Building Educational Applications",
    month = aug,
    year = "2019",
    address = "Florence, Italy",
    publisher = "Association for Computational Linguistics",
    url = "https://aclanthology.org/W19-4406/",
    doi = "10.18653/v1/W19-4406",
    pages = "52--75",
}

@inproceedings{ngetal2014conll,
    title = "The {C}o{NLL}-2014 Shared Task on Grammatical Error Correction",
    author = "Ng, Hwee Tou  and
      Wu, Siew Mei  and
      Briscoe, Ted  and
      Hadiwinoto, Christian  and
      Susanto, Raymond Hendy  and
      Bryant, Christopher",
    editor = "Ng, Hwee Tou  and
      Wu, Siew Mei  and
      Briscoe, Ted  and
      Hadiwinoto, Christian  and
      Susanto, Raymond Hendy  and
      Bryant, Christopher",
    booktitle = "Proceedings of the Eighteenth Conference on Computational Natural Language Learning: Shared Task",
    month = jun,
    year = "2014",
    address = "Baltimore, Maryland",
    publisher = "Association for Computational Linguistics",
    url = "https://aclanthology.org/W14-1701/",
    doi = "10.3115/v1/W14-1701",
    pages = "1--14"
}

@inproceedings{napoles2017jfleg,
    title = "{JFLEG}: A Fluency Corpus and Benchmark for Grammatical Error Correction",
    author = "Napoles, Courtney  and
      Sakaguchi, Keisuke  and
      Tetreault, Joel",
    editor = "Lapata, Mirella  and
      Blunsom, Phil  and
      Koller, Alexander",
    booktitle = "Proceedings of the 15th Conference of the {E}uropean Chapter of the Association for Computational Linguistics: Volume 2, Short Papers",
    month = apr,
    year = "2017",
    address = "Valencia, Spain",
    publisher = "Association for Computational Linguistics",
    url = "https://aclanthology.org/E17-2037/",
    pages = "229--234",
}

@misc{anthropic2025haiku45,
  title = {Claude {Haiku} 4.5 System Card},
  author       = {{Anthropic}},
  year         = {2025},
  howpublished = {\url{https://www.anthropic.com/claude-haiku-4-5-system-card}},
  note         = {Accessed: 2026-05-25}
}

@misc{xai2026grok43,
  title        = {Grok 4.3},
  author       = {{xAI}},
  year         = {2026},
  howpublished = {\url{https://docs.x.ai/developers/models/grok-4.3}},
  note         = {Accessed: 2026-05-25}
}

@inproceedings{bryant2017ERRANT,
    title = "Automatic Annotation and Evaluation of Error Types for Grammatical Error Correction",
    author = "Bryant, Christopher  and
      Felice, Mariano  and
      Briscoe, Ted",
    editor = "Barzilay, Regina  and
      Kan, Min-Yen",
    booktitle = "Proceedings of the 55th Annual Meeting of the Association for Computational Linguistics (Volume 1: Long Papers)",
    month = jul,
    year = "2017",
    address = "Vancouver, Canada",
    publisher = "Association for Computational Linguistics",
    url = "https://aclanthology.org/P17-1074/",
    doi = "10.18653/v1/P17-1074",
    pages = "793--805",
}

@inproceedings{
    hu2022lora,
    title={Lo{RA}: Low-Rank Adaptation of Large Language Models},
    author = {Hu, Edward J. and Shen, Yelong and Wallis, Phillip and Allen-Zhu, Zeyuan and Li, Yuanzhi and Wang, Shean and Wang, Lu and Chen, Weizhu},
    booktitle={International Conference on Learning Representations},
    year={2022},
    url={https://openreview.net/forum?id=nZeVKeeFYf9}
}

@inproceedings{
    he2023debertav,
    title={De{BERT}a{V3}: Improving De{BERT}a using {ELECTRA}-Style Pre-Training with Gradient-Disentangled Embedding Sharing},
    author={Pengcheng He and Jianfeng Gao and Weizhu Chen},
    booktitle={The Eleventh International Conference on Learning Representations },
    year={2023},
    url={https://openreview.net/forum?id=sE7-XhLxHA}
}

@inproceedings{kobayashi2024SEEDA,
    title = "Large Language Models Are State-of-the-Art Evaluator for Grammatical Error Correction",
    author = "Kobayashi, Masamune  and
      Mita, Masato  and
      Komachi, Mamoru",
    editor = {Kochmar, Ekaterina  and
      Bexte, Marie  and
      Burstein, Jill  and
      Horbach, Andrea  and
      Laarmann-Quante, Ronja  and
      Tack, Ana{\"i}s  and
      Yaneva, Victoria  and
      Yuan, Zheng},
    booktitle = "Proceedings of the 19th Workshop on Innovative Use of NLP for Building Educational Applications (BEA 2024)",
    month = jun,
    year = "2024",
    address = "Mexico City, Mexico",
    publisher = "Association for Computational Linguistics",
    url = "https://aclanthology.org/2024.bea-1.6/",
    pages = "68--77",
}

@inproceedings{gecdemo2025,
    title = "gec-metrics: A Unified Library for Grammatical Error Correction Evaluation",
    author = "Goto, Takumi  and
      Sakai, Yusuke  and
      Watanabe, Taro",
    editor = "Mishra, Pushkar  and
      Muresan, Smaranda  and
      Yu, Tao",
    booktitle = "Proceedings of the 63rd Annual Meeting of the Association for Computational Linguistics (Volume 3: System Demonstrations)",
    month = jul,
    year = "2025",
    address = "Vienna, Austria",
    publisher = "Association for Computational Linguistics",
    url = "https://aclanthology.org/2025.acl-demo.50/",
    doi = "10.18653/v1/2025.acl-demo.50",
    pages = "524--534",
    ISBN = "979-8-89176-253-4",
}

@inproceedings{gongetal2022,
    title = "Revisiting Grammatical Error Correction Evaluation and Beyond",
    author = "Gong, Peiyuan  and
      Liu, Xuebo  and
      Huang, Heyan  and
      Zhang, Min",
    editor = "Goldberg, Yoav  and
      Kozareva, Zornitsa  and
      Zhang, Yue",
    booktitle = "Proceedings of the 2022 Conference on Empirical Methods in Natural Language Processing",
    month = dec,
    year = "2022",
    address = "Abu Dhabi, United Arab Emirates",
    publisher = "Association for Computational Linguistics",
    url = "https://aclanthology.org/2022.emnlp-main.463/",
    doi = "10.18653/v1/2022.emnlp-main.463",
    pages = "6891--6902",
}

@inproceedings{
    Zhang2020BERTScore,
    title={{BERTScore}: Evaluating Text Generation with BERT},
    author = {Zhang, Tianyi and Kishore, Varsha and Wu, Felix and Weinberger, Kilian Q. and Artzi, Yoav},
    booktitle={International Conference on Learning Representations},
    year={2020},
    url={https://openreview.net/forum?id=SkeHuCVFDr}
}

@inproceedings{gotouetal2020taking,
    title = "Taking the Correction Difficulty into Account in Grammatical Error Correction Evaluation",
    author = "Gotou, Takumi  and
      Nagata, Ryo  and
      Mita, Masato  and
      Hanawa, Kazuaki",
    editor = "Scott, Donia  and
      Bel, Nuria  and
      Zong, Chengqing",
    booktitle = "Proceedings of the 28th International Conference on Computational Linguistics",
    month = dec,
    year = "2020",
    address = "Barcelona, Spain (Online)",
    publisher = "International Committee on Computational Linguistics",
    url = "https://aclanthology.org/2020.coling-main.188/",
    doi = "10.18653/v1/2020.coling-main.188",
    pages = "2085--2095",
}

@article{tacl_a_00282,
    author = {Napoles, Courtney and Nădejde, Maria and Tetreault, Joel},
    title = {Enabling Robust Grammatical Error Correction in New Domains: Data Sets, Metrics, and Analyses},
    journal = {Transactions of the Association for Computational Linguistics},
    volume = {7},
    pages = {551-566},
    year = {2019},
    month = {09},
    issn = {2307-387X},
    doi = {10.1162/tacl_a_00282},
    url = {https://doi.org/10.1162/tacl_a_00282},
    eprint = {https://direct.mit.edu/tacl/article-pdf/doi/10.1162/tacl_a_00282/1923556/tacl_a_00282.pdf},
}

@inproceedings{napolesetal2015ground,
    title = "Ground Truth for Grammatical Error Correction Metrics",
    author = "Napoles, Courtney  and
      Sakaguchi, Keisuke  and
      Post, Matt  and
      Tetreault, Joel",
    editor = "Zong, Chengqing  and
      Strube, Michael",
    booktitle = "Proceedings of the 53rd Annual Meeting of the Association for Computational Linguistics and the 7th International Joint Conference on Natural Language Processing (Volume 2: Short Papers)",
    month = jul,
    year = "2015",
    address = "Beijing, China",
    publisher = "Association for Computational Linguistics",
    url = "https://aclanthology.org/P15-2097/",
    doi = "10.3115/v1/P15-2097",
    pages = "588--593"
}

@inproceedings{koyamaetal2024,
    title = "n-gram {F}-score for Evaluating Grammatical Error Correction",
    author = "Koyama, Shota  and
      Nagata, Ryo  and
      Takamura, Hiroya  and
      Okazaki, Naoaki",
    editor = "Mahamood, Saad  and
      Minh, Nguyen Le  and
      Ippolito, Daphne",
    booktitle = "Proceedings of the 17th International Natural Language Generation Conference",
    month = sep,
    year = "2024",
    address = "Tokyo, Japan",
    publisher = "Association for Computational Linguistics",
    url = "https://aclanthology.org/2024.inlg-main.25/",
    doi = "10.18653/v1/2024.inlg-main.25",
    pages = "303--313",
}

@inproceedings{some2020,
    title = "{SOME}: Reference-less Sub-Metrics Optimized for Manual Evaluations of Grammatical Error Correction",
    author = "Yoshimura, Ryoma and Kaneko, Masahiro and Kajiwara, Tomoyuki and Komachi, Mamoru",
    booktitle = "Proceedings of the 28th International Conference on Computational Linguistics",
    month = dec,
    year = "2020",
    address = "Barcelona, Spain (Online)",
    publisher = "International Committee on Computational Linguistics",
    url = "https://aclanthology.org/2020.coling-main.573/",
    doi = "10.18653/v1/2020.coling-main.573",
    pages = "6516--6522",
}

@inproceedings{islammagnani2021end,
    title = "Is this the end of the gold standard? A straightforward reference-less grammatical error correction metric",
    author = "Islam, Md Asadul  and
      Magnani, Enrico",
    editor = "Moens, Marie-Francine  and
      Huang, Xuanjing  and
      Specia, Lucia  and
      Yih, Scott Wen-tau",
    booktitle = "Proceedings of the 2021 Conference on Empirical Methods in Natural Language Processing",
    month = nov,
    year = "2021",
    address = "Online and Punta Cana, Dominican Republic",
    publisher = "Association for Computational Linguistics",
    url = "https://aclanthology.org/2021.emnlp-main.239/",
    doi = "10.18653/v1/2021.emnlp-main.239",
    pages = "3009--3015",
}

@inproceedings{maedaetal2022impara,
    title = "{IMPARA}: Impact-Based Metric for {GEC} Using Parallel Data",
    author = "Maeda, Koki  and
      Kaneko, Masahiro  and
      Okazaki, Naoaki",
    editor = "Calzolari, Nicoletta  and
      Huang, Chu-Ren  and
      Kim, Hansaem  and
      Pustejovsky, James  and
      Wanner, Leo  and
      Choi, Key-Sun  and
      Ryu, Pum-Mo  and
      Chen, Hsin-Hsi  and
      Donatelli, Lucia  and
      Ji, Heng  and
      Kurohashi, Sadao  and
      Paggio, Patrizia  and
      Xue, Nianwen  and
      Kim, Seokhwan  and
      Hahm, Younggyun  and
      He, Zhong  and
      Lee, Tony Kyungil  and
      Santus, Enrico  and
      Bond, Francis  and
      Na, Seung-Hoon",
    booktitle = "Proceedings of the 29th International Conference on Computational Linguistics",
    month = oct,
    year = "2022",
    address = "Gyeongju, Republic of Korea",
    publisher = "International Committee on Computational Linguistics",
    url = "https://aclanthology.org/2022.coling-1.316/",
    pages = "3578--3588",
}

@inproceedings{yeetal2025cleme2,
    title = "{CLEME}2.0: Towards Interpretable Evaluation by Disentangling Edits for Grammatical Error Correction",
    author = "Ye, Jingheng  and
      Xu, Zishan  and
      Li, Yinghui  and
      Song, Linlin  and
      Zhou, Qingyu  and
      Zheng, Hai-Tao  and
      Shen, Ying  and
      Jiang, Wenhao  and
      Kim, Hong-Gee  and
      Liu, Ruitong  and
      Su, Xin  and
      Shan, Zifei",
    editor = "Che, Wanxiang  and
      Nabende, Joyce  and
      Shutova, Ekaterina  and
      Pilehvar, Mohammad Taher",
    booktitle = "Proceedings of the 63rd Annual Meeting of the Association for Computational Linguistics (Volume 1: Long Papers)",
    month = jul,
    year = "2025",
    address = "Vienna, Austria",
    publisher = "Association for Computational Linguistics",
    url = "https://aclanthology.org/2025.acl-long.10/",
    doi = "10.18653/v1/2025.acl-long.10",
    pages = "204--222",
    ISBN = "979-8-89176-251-0",
}

@inproceedings{zhan2026jelv,
  title = {{JELV}: A Judge of Edit-Level Validity for Evaluation and Automated Reference Expansion in Grammatical Error Correction},
  author = {Zhan, Yuhao and Zhang, Yuqing and Yuan, Jing and Ma, Qixiang and Yang, Zhiqi and Gu, Yu and Liu, Zemin and Wu, Fei},
  booktitle = {Proceedings of the AAAI Conference on Artificial Intelligence},
  volume = {40},
  number = {41},
  pages = {34611--34619},
  year = {2026},
  doi = {10.1609/aaai.v40i41.40761},
  url = {https://ojs.aaai.org/index.php/AAAI/article/view/40761}
}

@inproceedings{dahlmeier2012,
    title = "Better Evaluation for Grammatical Error Correction",
    author = "Dahlmeier, Daniel  and
      Ng, Hwee Tou",
    editor = "Fosler-Lussier, Eric  and
      Riloff, Ellen  and
      Bangalore, Srinivas",
    booktitle = "Proceedings of the 2012 Conference of the North {A}merican Chapter of the Association for Computational Linguistics: Human Language Technologies",
    month = jun,
    year = "2012",
    address = "Montr{\'e}al, Canada",
    publisher = "Association for Computational Linguistics",
    url = "https://aclanthology.org/N12-1067/",
    pages = "568--572"
}

@inproceedings{chollampatt_2018_reassessment,
    title = "A Reassessment of Reference-Based Grammatical Error Correction Metrics",
    author = "Chollampatt, Shamil  and
      Ng, Hwee Tou",
    editor = "Bender, Emily M.  and
      Derczynski, Leon  and
      Isabelle, Pierre",
    booktitle = "Proceedings of the 27th International Conference on Computational Linguistics",
    month = aug,
    year = "2018",
    address = "Santa Fe, New Mexico, USA",
    publisher = "Association for Computational Linguistics",
    url = "https://aclanthology.org/C18-1231/",
    pages = "2730--2741",
}

@inproceedings{Christiano2017,
  author = {Christiano, Paul F. and Leike, Jan and Brown, Tom B. and Martic, Miljan and Legg, Shane and Amodei, Dario},
  title = {Deep Reinforcement Learning from Human Preferences},
  booktitle = {Advances in Neural Information Processing Systems},
  year = {2017},
  volume = {30},
  pages = {4302--4310},
  url = {https://proceedings.neurips.cc/paper/2017/hash/d5e2c0adad503c91f91df240d0cd4e49-Abstract.html}
}

@misc{ziegler2020,
      title={Fine-Tuning Language Models from Human Preferences}, 
      author={Daniel M. Ziegler and Nisan Stiennon and Jeffrey Wu and Tom B. Brown and Alec Radford and Dario Amodei and Paul Christiano and Geoffrey Irving},
      year={2020},
      eprint={1909.08593},
      archivePrefix={arXiv},
      primaryClass={cs.CL},
      url={https://arxiv.org/abs/1909.08593}, 
}

@inproceedings{Stiennon2020,
  author = {Stiennon, Nisan and Ouyang, Long and Wu, Jeffrey and Ziegler, Daniel M. and Lowe, Ryan and Voss, Chelsea and Radford, Alec and Amodei, Dario and Christiano, Paul F.},
  title = {Learning to Summarize with Human Feedback},
  booktitle = {Advances in Neural Information Processing Systems},
  volume = {33},
  pages = {3008--3021},
  year = {2020},
  url = {https://proceedings.neurips.cc/paper/2020/hash/1f89885d556929e98d3ef9b86448f951-Abstract.html}
}

@inproceedings{ouyang2022training,
  author = {Ouyang, Long and Wu, Jeff and Jiang, Xu and Almeida, Diogo and Wainwright, Carroll L. and Mishkin, Pamela and Zhang, Chong and Agarwal, Sandhini and Slama, Katarina and Ray, Alex and Schulman, John and Hilton, Jacob and Kelton, Fraser and Miller, Luke and Simens, Maddie and Askell, Amanda and Welinder, Peter and Christiano, Paul and Leike, Jan and Lowe, Ryan},
  title = {Training Language Models to Follow Instructions with Human Feedback},
  booktitle = {Advances in Neural Information Processing Systems},
  volume = {35},
  pages = {27730--27744},
  year = {2022},
  url = {https://proceedings.neurips.cc/paper_files/paper/2022/hash/b1efde53be364a73914f58805a001731-Abstract.html}
}

@misc{bai2022traininghelpfulharmlessassistant,
      title={Training a Helpful and Harmless Assistant with Reinforcement Learning from Human Feedback}, 
      author={Yuntao Bai and Andy Jones and Kamal Ndousse and Amanda Askell and Anna Chen and Nova DasSarma and Dawn Drain and Stanislav Fort and Deep Ganguli and Tom Henighan and Nicholas Joseph and Saurav Kadavath and Jackson Kernion and Tom Conerly and Sheer El-Showk and Nelson Elhage and Zac Hatfield-Dodds and Danny Hernandez and Tristan Hume and Scott Johnston and Shauna Kravec and Liane Lovitt and Neel Nanda and Catherine Olsson and Dario Amodei and Tom Brown and Jack Clark and Sam McCandlish and Chris Olah and Ben Mann and Jared Kaplan},
      year={2022},
      eprint={2204.05862},
      archivePrefix={arXiv},
      primaryClass={cs.CL},
      url={https://arxiv.org/abs/2204.05862}, 
}

@inproceedings{
rafailov2023direct,
title={Direct Preference Optimization: Your Language Model is Secretly a Reward Model},
author={Rafael Rafailov and Archit Sharma and Eric Mitchell and Christopher D Manning and Stefano Ermon and Chelsea Finn},
booktitle={Thirty-seventh Conference on Neural Information Processing Systems},
year={2023},
url={https://openreview.net/forum?id=HPuSIXJaa9}
}

@inproceedings{sellametal2020bleurt,
    title = "{BLEURT}: Learning Robust Metrics for Text Generation",
    author = "Sellam, Thibault  and
      Das, Dipanjan  and
      Parikh, Ankur",
    editor = "Jurafsky, Dan  and
      Chai, Joyce  and
      Schluter, Natalie  and
      Tetreault, Joel",
    booktitle = "Proceedings of the 58th Annual Meeting of the Association for Computational Linguistics",
    month = jul,
    year = "2020",
    address = "Online",
    publisher = "Association for Computational Linguistics",
    url = "https://aclanthology.org/2020.acl-main.704/",
    doi = "10.18653/v1/2020.acl-main.704",
    pages = "7881--7892",
}

@inproceedings{rei2020comet,
    title = "{COMET}: A Neural Framework for {MT} Evaluation",
    author = "Rei, Ricardo  and
      Stewart, Craig  and
      Farinha, Ana C  and
      Lavie, Alon",
    editor = "Webber, Bonnie  and
      Cohn, Trevor  and
      He, Yulan  and
      Liu, Yang",
    booktitle = "Proceedings of the 2020 Conference on Empirical Methods in Natural Language Processing (EMNLP)",
    month = nov,
    year = "2020",
    address = "Online",
    publisher = "Association for Computational Linguistics",
    url = "https://aclanthology.org/2020.emnlp-main.213/",
    doi = "10.18653/v1/2020.emnlp-main.213",
    pages = "2685--2702",
}

@inproceedings{NEURIPS2021_e4d2b6e6,
  author = {Yuan, Weizhe and Neubig, Graham and Liu, Pengfei},
  title = {{BARTScore}: Evaluating Generated Text as Text Generation},
  booktitle = {Advances in Neural Information Processing Systems},
  volume = {34},
  pages = {27263--27277},
  year = {2021},
  url = {https://proceedings.neurips.cc/paper/2021/hash/e4d2b6e6fdeca3e60e0f1a62fee3d9dd-Abstract.html}
}

@inproceedings{zhong2022towards,
    title = "Towards a Unified Multi-Dimensional Evaluator for Text Generation",
    author = "Zhong, Ming  and
      Liu, Yang  and
      Yin, Da  and
      Mao, Yuning  and
      Jiao, Yizhu  and
      Liu, Pengfei  and
      Zhu, Chenguang  and
      Ji, Heng  and
      Han, Jiawei",
    editor = "Goldberg, Yoav  and
      Kozareva, Zornitsa  and
      Zhang, Yue",
    booktitle = "Proceedings of the 2022 Conference on Empirical Methods in Natural Language Processing",
    month = dec,
    year = "2022",
    address = "Abu Dhabi, United Arab Emirates",
    publisher = "Association for Computational Linguistics",
    url = "https://aclanthology.org/2022.emnlp-main.131/",
    doi = "10.18653/v1/2022.emnlp-main.131",
    pages = "2023--2038",
}

@inproceedings{fu2023gptscore,
    title = "{GPTS}core: Evaluate as You Desire",
    author = "Fu, Jinlan  and
      Ng, See-Kiong  and
      Jiang, Zhengbao  and
      Liu, Pengfei",
    editor = "Duh, Kevin  and
      Gomez, Helena  and
      Bethard, Steven",
    booktitle = "Proceedings of the 2024 Conference of the North American Chapter of the Association for Computational Linguistics: Human Language Technologies (Volume 1: Long Papers)",
    month = jun,
    year = "2024",
    address = "Mexico City, Mexico",
    publisher = "Association for Computational Linguistics",
    url = "https://aclanthology.org/2024.naacl-long.365/",
    doi = "10.18653/v1/2024.naacl-long.365",
    pages = "6556--6576",
}

@inproceedings{liu2023geval,
    title = "{G}-Eval: {NLG} Evaluation using {Gpt-4} with Better Human Alignment",
    author = "Liu, Yang  and
      Iter, Dan  and
      Xu, Yichong  and
      Wang, Shuohang  and
      Xu, Ruochen  and
      Zhu, Chenguang",
    editor = "Bouamor, Houda  and
      Pino, Juan  and
      Bali, Kalika",
    booktitle = "Proceedings of the 2023 Conference on Empirical Methods in Natural Language Processing",
    month = dec,
    year = "2023",
    address = "Singapore",
    publisher = "Association for Computational Linguistics",
    url = "https://aclanthology.org/2023.emnlp-main.153/",
    doi = "10.18653/v1/2023.emnlp-main.153",
    pages = "2511--2522",
}

@inproceedings{wang2023chatgpt,
    title = "Is {C}hat{GPT} a Good {NLG} Evaluator? A Preliminary Study",
    author = "Wang, Jiaan  and
      Liang, Yunlong  and
      Meng, Fandong  and
      Sun, Zengkui  and
      Shi, Haoxiang  and
      Li, Zhixu  and
      Xu, Jinan  and
      Qu, Jianfeng  and
      Zhou, Jie",
    editor = "Dong, Yue  and
      Xiao, Wen  and
      Wang, Lu  and
      Liu, Fei  and
      Carenini, Giuseppe",
    booktitle = "Proceedings of the 4th New Frontiers in Summarization Workshop",
    month = dec,
    year = "2023",
    address = "Singapore",
    publisher = "Association for Computational Linguistics",
    url = "https://aclanthology.org/2023.newsum-1.1/",
    doi = "10.18653/v1/2023.newsum-1.1",
    pages = "1--11",
}

@inproceedings{zheng2023judging,
  author = {Zheng, Lianmin and Chiang, Wei-Lin and Sheng, Ying and Zhuang, Siyuan and Wu, Zhanghao and Zhuang, Yonghao and Lin, Zi and Li, Zhuohan and Li, Dacheng and Xing, Eric P. and Zhang, Hao and Gonzalez, Joseph E. and Stoica, Ion},
  title = {Judging {LLM}-as-a-{Judge} with {MT-Bench} and {Chatbot Arena}},
  booktitle = {Advances in Neural Information Processing Systems},
  volume = {36},
  year = {2023},
  pages = {46595--46623},
  url = {https://proceedings.neurips.cc/paper_files/paper/2023/hash/91f18a1287b398d378ef22505bf41832-Abstract-Datasets_and_Benchmarks.html}
}

@inproceedings{
chen2024alpagasus,
title={{AlpaGasus}: Training a Better Alpaca with Fewer Data},
author={Lichang Chen and Shiyang Li and Jun Yan and Hai Wang and Kalpa Gunaratna and Vikas Yadav and Zheng Tang and Vijay Srinivasan and Tianyi Zhou and Heng Huang and Hongxia Jin},
booktitle={The Twelfth International Conference on Learning Representations},
year={2024},
url={https://openreview.net/forum?id=FdVXgSJhvz}
}

@inproceedings{dubois2024lengthcontrolled,
  author = {Dubois, Yann and Liang, Percy and Hashimoto, Tatsunori B.},
  title = {Length-Controlled {AlpacaEval}: A Simple Debiasing of Automatic Evaluators},
  booktitle = {First Conference on Language Modeling},
  year = {2024},
  url = {https://openreview.net/forum?id=CybBmzWBX0}
}

@inproceedings{wang2024rewardbench,
    title = "{R}eward{B}ench: Evaluating Reward Models for Language Modeling",
    author = "Lambert, Nathan  and
      Pyatkin, Valentina  and
      Morrison, Jacob  and
      Miranda, LJ  and
      Lin, Bill Yuchen  and
      Chandu, Khyathi  and
      Dziri, Nouha  and
      Kumar, Sachin  and
      Zick, Tom  and
      Choi, Yejin  and
      Smith, Noah A.  and
      Hajishirzi, Hannaneh",
    editor = "Chiruzzo, Luis  and
      Ritter, Alan  and
      Wang, Lu",
    booktitle = "Findings of the Association for Computational Linguistics: NAACL 2025",
    month = apr,
    year = "2025",
    address = "Albuquerque, New Mexico",
    publisher = "Association for Computational Linguistics",
    url = "https://aclanthology.org/2025.findings-naacl.96/",
    doi = "10.18653/v1/2025.findings-naacl.96",
    pages = "1755--1797",
    ISBN = "979-8-89176-195-7",
}

@inproceedings{asanoetal2017reference,
    title = "Reference-based Metrics can be Replaced with Reference-less Metrics in Evaluating Grammatical Error Correction Systems",
    author = "Asano, Hiroki  and
      Mizumoto, Tomoya  and
      Inui, Kentaro",
    editor = "Kondrak, Greg  and
      Watanabe, Taro",
    booktitle = "Proceedings of the Eighth International Joint Conference on Natural Language Processing (Volume 2: Short Papers)",
    month = nov,
    year = "2017",
    address = "Taipei, Taiwan",
    publisher = "Asian Federation of Natural Language Processing",
    url = "https://aclanthology.org/I17-2058/",
    pages = "343--348",
}

@inproceedings{choshenabend2018reference,
    title = "Reference-less Measure of Faithfulness for Grammatical Error Correction",
    author = "Choshen, Leshem  and
      Abend, Omri",
    editor = "Walker, Marilyn  and
      Ji, Heng  and
      Stent, Amanda",
    booktitle = "Proceedings of the 2018 Conference of the North {A}merican Chapter of the Association for Computational Linguistics: Human Language Technologies, Volume 2 (Short Papers)",
    month = jun,
    year = "2018",
    address = "New Orleans, Louisiana",
    publisher = "Association for Computational Linguistics",
    url = "https://aclanthology.org/N18-2020/",
    doi = "10.18653/v1/N18-2020",
    pages = "124--129",
}

@misc{fang2023,
      title={Is {ChatGPT} a Highly Fluent Grammatical Error Correction System? A Comprehensive Evaluation}, 
      author={Tao Fang and Shu Yang and Kaixin Lan and Derek F. Wong and Jinpeng Hu and Lidia S. Chao and Yue Zhang},
      year={2023},
      eprint={2304.01746},
      archivePrefix={arXiv},
      primaryClass={cs.CL},
      url={https://arxiv.org/abs/2304.01746}, 
}

@inproceedings{katinskaia2024,
    title = "{GPT}-3.5 for Grammatical Error Correction",
    author = "Katinskaia, Anisia  and
      Yangarber, Roman",
    editor = "Calzolari, Nicoletta  and
      Kan, Min-Yen  and
      Hoste, Veronique  and
      Lenci, Alessandro  and
      Sakti, Sakriani  and
      Xue, Nianwen",
    booktitle = "Proceedings of the 2024 Joint International Conference on Computational Linguistics, Language Resources and Evaluation (LREC-COLING 2024)",
    month = may,
    year = "2024",
    address = "Torino, Italia",
    publisher = "ELRA and ICCL",
    url = "https://aclanthology.org/2024.lrec-main.692/",
    pages = "7831--7843",
}

@article{kobayashi2024revisiting,
    author = {Kobayashi, Masamune and Mita, Masato and Komachi, Mamoru},
    title = {Revisiting Meta-evaluation for Grammatical Error Correction},
    journal = {Transactions of the Association for Computational Linguistics},
    volume = {12},
    pages = {837-855},
    year = {2024},
    month = {07},
    issn = {2307-387X},
    doi = {10.1162/tacl_a_00676},
    url = {https://doi.org/10.1162/tacl_a_00676},
    eprint = {https://direct.mit.edu/tacl/article-pdf/doi/10.1162/tacl_a_00676/2461947/tacl_a_00676.pdf},
}

@misc{openai2024gpt4o,
  title        = {{GPT-4o} System Card},
  author       = {{OpenAI}},
  year         = {2024},
  eprint       = {2410.21276},
  archivePrefix = {arXiv},
  url          = {https://arxiv.org/abs/2410.21276}
}

@misc{openai2025gpt41,
  author       = {{OpenAI}},
  title        = {Introducing {GPT-4.1} in the {API}},
  year         = {2025},
  howpublished = {\url{https://openai.com/index/gpt-4-1/}}
}

@misc{openai2024gpt4technicalreport,
  author        = {{OpenAI}},
  title         = {{GPT-4} Technical Report},
  year          = {2023},
  eprint        = {2303.08774},
  archivePrefix = {arXiv},
  primaryClass  = {cs.CL}
}

\clearpage
\appendix
\section{Dataset Statistics} \label{app:data_statistics}
\paragraph{Dataset Composition.}
The final dataset contains 2,400 preference pairs from 1,320 unique source sentences, drawn from BEA-2019 and JFLEG (Table~\ref{tab:dataset_composition}).
CoNLL-2014-derived data are excluded to avoid source overlap with SEEDA.

\begin{table}[h]
    \centering
    \small
    \begin{tabular}{lcc}
    \toprule
    Dataset & Preference pairs & Unique sources \\
    \midrule
    BEA-2019 & 1,500 & 881 \\
    JFLEG & 900 & 439 \\
    \midrule
    Total & 2,400 & 1,320 \\
    \bottomrule
    \end{tabular}
    \caption{Dataset composition by source benchmark.}
    \label{tab:dataset_composition}
\end{table}

\paragraph{Preference Filtering.}
The three LLM judges evaluated 8,093 candidate pairs, of which 4,846 (59.9\%)
reached unanimous agreement on the overall preference, while 3,247 received
2--1 split judgments. 
From these unanimous pairs, we randomly sampled 2,400 pairs to construct the final training set.

\paragraph{Candidate Source Coverage.}
GPT-4o-generated corrections appear in 87.0\% of all preference pairs, accounting for 62.5\% of preferred candidates and 40.4\% of dispreferred candidates (Table~\ref{tab:gpt4o_coverage}). This indicates that the training data provides substantial supervision over LLM-generated corrections, including rewrite-style corrections. The minimal-edit GPT-4o variant has a win rate of 53.8\%, while the rewrite-oriented GPT-4o variant has a higher win rate of 69.6\%. This difference suggests that rewrite-oriented corrections are frequently preferred when they provide valid and fluent alternatives, which we further examine in Section~\ref{sec:preference}.  

\begin{table}[h]
    \centering
    \small
    \resizebox{\columnwidth}{!}{
    \begin{tabular}{lr}
    \toprule
    Statistic & Value \\
    \midrule
    Pairs containing at least one GPT-4o correction & 87.0\% \\
    Preferred candidates from GPT-4o corrections & 62.5\% \\
    Dispreferred candidates from GPT-4o corrections & 40.4\% \\
    Minimal-edit GPT-4o variant win rate & 53.8\% \\
    Rewrite-oriented GPT-4o variant win rate & 69.6\% \\
    \bottomrule
    \end{tabular}
    }
    \caption{Coverage and preference rates of GPT-4o-generated corrections.}
    \label{tab:gpt4o_coverage}
\end{table}

\paragraph{Style Pair Distribution.}
Table~\ref{tab:style_pairs} reports the distribution of preference pairs
according to the correction styles of the preferred and dispreferred candidates.
The arrow denotes the preference direction, with the style on the left
corresponding to the preferred correction and the style on the right
to the dispreferred correction.
The dataset contains substantial comparisons between rewrite-style and
minimal-edit corrections, as well as comparisons involving fluency-oriented
and rewrite-style corrections.

\begin{table}[h]
    \centering
    \small
    \begin{tabular}{lr}
    \toprule
    Style pair ($y^{+} \rightarrow y^{-}$) & Count \\
    \midrule
    rewrite $\rightarrow$ minimal edit & 601 \\
    minimal edit $\rightarrow$ rewrite & 560 \\
    rewrite $\rightarrow$ fluency edit & 518 \\
    rewrite $\rightarrow$ rewrite & 382 \\
    fluency edit $\rightarrow$ fluency edit & 311 \\
    fluency edit $\rightarrow$ rewrite & 28 \\
    \midrule
    Total & 2,400 \\
    \bottomrule
    \end{tabular}
    \caption{Distribution of preference pairs by correction style.}
    \label{tab:style_pairs}
\end{table}

\paragraph{Criteria Labels and Instance Format.}
For each criterion, Table~\ref{tab:criteria_distribution} reports whether the overall preferred correction $y^{+}$ or the dispreferred correction $y^{-}$ is preferred under that criterion. Each training instance contains a source sentence, ERRANT-identified error spans, a preferred correction $y^{+}$, a dispreferred correction $y^{-}$, criteria-level preferences, span-resolution labels for both candidates, and metadata.  

\begin{table}[h]
    \centering
    \small
    \begin{tabular}{lcc}
    \toprule
    Criterion & $y^{+}$ preferred & $y^{-}$ preferred \\
    \midrule
    Grammaticality & 2,367 & 33 \\
    Faithfulness & 1,264 & 1,136 \\
    Fluency & 2,179 & 221 \\
    \bottomrule
    \end{tabular}
    \caption{Alignment between criteria-level preferences and the overall preference.}
    \label{tab:criteria_distribution}
\end{table}

\begin{table*}[t]
    \centering
    \small
    \setlength{\tabcolsep}{3pt}

    \resizebox{\textwidth}{!}{%
    \begin{tabular}{@{}lrrrr lrrrr@{}}
    \toprule
    \multicolumn{5}{c}{SEEDA-E} &
    \multicolumn{5}{c}{SEEDA-S} \\
    \cmidrule(lr){1-5} \cmidrule(lr){6-10}
    System & Human & ERRANT ($\Delta$) & SOME ($\Delta$) & SURE ($\Delta$) &
    System & Human & ERRANT ($\Delta$) & SOME ($\Delta$) & SURE ($\Delta$) \\
    \midrule
    REF-F         & 1  & 14 (+13) & 1 (+0)  & 1 (+0)  &
    REF-F         & 1  & 14 (+13) & 1 (+0)  & 1 (+0)  \\

    GPT-3.5       & 2  & 12 (+10) & 2 (+0)  & 2 (+0)  &
    GPT-3.5       & 2  & 12 (+10) & 2 (+0)  & 2 (+0)  \\

    TransGEC      & 3  & 5 (+2)   & 4 (+1)  & 3 (+0)  &
    T5            & 3  & 6 (+3)   & 3 (+0)  & 4 (+1)  \\

    T5            & 4  & 6 (+2)   & 3 (-1)  & 4 (+0)  &
    TransGEC      & 4  & 5 (+1)   & 4 (+0)  & 3 (-1)  \\

    REF-M         & 5  & 13 (+8)  & 5 (+0)  & 5 (+0)  &
    REF-M         & 5  & 13 (+8)  & 5 (+0)  & 5 (+0)  \\

    Riken-Tohoku  & 6  & 1 (-5)   & 8 (+2)  & 7 (+1)  &
    BERT-fuse     & 6  & 3 (-3)   & 6 (+0)  & 6 (+0)  \\

    BERT-fuse     & 7  & 3 (-4)   & 6 (-1)  & 6 (-1)  &
    Riken-Tohoku  & 7  & 1 (-6)   & 8 (+1)  & 7 (+0)  \\

    UEDIN-MS      & 8  & 2 (-6)   & 7 (-1)  & 8 (+0)  &
    PIE           & 8  & 8 (+0)   & 9 (+1)  & 9 (+1)  \\

    PIE           & 9  & 8 (-1)   & 9 (+0)  & 9 (+0)  &
    LM-Critic     & 9  & 9 (+0)   & 11 (+2) & 11 (+2) \\

    GECToR-BERT   & 10 & 7 (-3)   & 10 (+0) & 10 (+0) &
    TemplateGEC   & 10 & 10 (+0)  & 12 (+2) & 12 (+2) \\

    LM-Critic     & 11 & 9 (-2)   & 11 (+0) & 11 (+0) &
    GECToR-BERT   & 11 & 7 (-4)   & 10 (-1) & 10 (-1) \\

    GECToR-ens    & 12 & 4 (-8)   & 14 (+2) & 14 (+2) &
    UEDIN-MS      & 12 & 2 (-10)  & 7 (-5)  & 8 (-4)  \\

    TemplateGEC   & 13 & 10 (-3)  & 12 (-1) & 12 (-1) &
    GECToR-ens    & 13 & 4 (-9)   & 14 (+1) & 14 (+1) \\

    BART          & 14 & 11 (-3)  & 13 (-1) & 13 (-1) &
    BART          & 14 & 11 (-3)  & 13 (-1) & 13 (-1) \\
    \bottomrule
    \end{tabular}%
    }

    \caption{
    System rankings under the SEEDA $+$Fluency setting.
    Human denotes the ranking induced by human judgments.
    $\Delta$ denotes the displacement from the human rank
    (metric rank minus human rank).
    }
    \label{tab:system_ranking}
\end{table*}
\begin{table*}[t]
    \centering
    \small
    \setlength{\tabcolsep}{4pt}
    \begin{tabular}{lcccccc}
    \toprule
    Metric
    & \shortstack{FCE Sys.\\($r/\rho$)}
    & \shortstack{Wiki Sys.\\($r/\rho$)}
    & \shortstack{Sys.\\Avg.}
    & \shortstack{FCE Sent.\\($r/\rho$)}
    & \shortstack{Wiki Sent.\\($r/\rho$)}
    & \shortstack{Sent.\\Avg.} \\
    \midrule
    SOME
    & 0.941/0.929
    & 0.923/\textbf{0.833}
    & 0.932/\textbf{0.881}
    & \textbf{0.653/0.643}
    & 0.402/0.408
    & \textbf{0.528/0.526} \\

    IMPARA
    & 0.940/\textbf{0.952}
    & \textbf{0.954}/0.786
    & \textbf{0.947}/0.869
    & 0.392/0.336
    & 0.347/0.339
    & 0.369/0.337 \\

    Scribendi
    & 0.777/0.905
    & 0.846/0.571
    & 0.812/0.738
    & 0.005/-0.039
    & 0.210/0.180
    & 0.108/0.070 \\

    SURE
    & \textbf{0.961}/0.929
    & 0.694/0.595
    & 0.827/0.762
    & 0.575/0.553
    & \textbf{0.425/0.435}
    & 0.500/0.494 \\
    \bottomrule
    \end{tabular}
    \caption{
    Zero-shot cross-domain evaluation on the FCE and Wiki domains of GMEG.
    System-level and sentence-level results are reported using Pearson's $r$
    and Spearman's $\rho$.
    SURE is evaluated without additional fine-tuning.
    The best score in each column and correlation measure is shown in bold.
    }
    \label{tab:gmeg_transfer}
\end{table*}

\begin{table}[t]
\centering
\small
\setlength{\tabcolsep}{8pt}
\renewcommand{\arraystretch}{1.00}

\resizebox{\columnwidth}{!}{%
\begin{tabular}{lcccc}
\toprule
\multirow{2}{*}{\raisebox{-0.58ex}{\textbf{Criteria}}}
& \multicolumn{2}{c}{All Pairs}
& \multicolumn{2}{c}{\textbf{$|\Delta h| \geq 1.0$}} \\
\cmidrule(lr){2-3}
\cmidrule(lr){4-5}
& SOME & SURE
& SOME & SURE \\
\midrule
Grammaticality & 0.738 & 0.680 & 0.944 & 0.852 \\
Fluency        & 0.698 & 0.669 & 0.911 & 0.871 \\
Faithfulness   & 0.841 & 0.538 & 0.988 & 0.734 \\
\bottomrule
\end{tabular}%
}

\caption{Criteria-level pairwise accuracy on TMU-GFM. SOME is an in-domain model trained on TMU-GFM, while SURE is evaluated zero-shot. All pairs includes all candidate pairs with different human scores, and $|\Delta h| \geq 1.0$ keeps only confident pairs whose human score difference is at least 1.0. TMU-GFM's meaning-preservation axis is mapped to faithfulness for comparison.}
\label{tab:tmu_gfm_axis}
\end{table}

%TMU-GFM에서 criteria-level pairwise accuracy를 평가한 결과이다. SOME은 TMU-GFM에서 학습된 in-domain model이고, SURE는 zero-shot으로 평가된다. All pairs는 human score가 다른 모든 candidate pair를 포함하며, ∣Δh∣≥1.0은 human score 차이가 1.0 이상인 confident pair만 포함한다. 비교를 위해 TMU-GFM의 meaning-preservation axis는 faithfulness에 대응시킨다.

\section{System-Ranking Analysis under $+$Fluency}
\label{sec:system_ranking}

To examine whether the evaluation mismatch observed for rewrite-style
corrections affects practical system-level conclusions, we compare complete
system rankings under the SEEDA $+$Fluency setting.
Table~\ref{tab:system_ranking} reports the rankings induced by human judgments,
ERRANT, SOME, and SURE.

The largest discrepancies for ERRANT occur for fluent rewrite-style systems.
In both SEEDA-E and SEEDA-S, REF-F and GPT-3.5 are ranked first and second
by humans but are placed 14th and 12th by ERRANT, respectively.
SOME and SURE substantially reduce these rank displacements and recover the
human top-two ordering.

\section{Additional Analyses} \label{app:criteria_analysis}
\subsection{Cross-Domain Transfer on GMEG} \label{app:gmeg_transfer}

To assess cross-domain generalization, we evaluate SURE on the FCE and Wiki domains of GMEG~\citep{tacl_a_00282} without additional fine-tuning.
We compare SURE with the reference-free baselines SOME, IMPARA, and Scribendi, and report Pearson's $r$ and Spearman's $\rho$ at both the system and sentence levels.
As shown in Table~\ref{tab:gmeg_transfer}, SURE shows competitive performance on FCE and stronger sentence-level results on Wiki, while its Wiki system-level correlations remain lower than those of SOME and IMPARA.
These results suggest reasonable cross-domain transfer at the sentence level, while system-level calibration remains sensitive to domain shift.

\subsection{Criteria-Level Transfer on TMU-GFM} \label{app:tmu_gfm}
Table~\ref{tab:tmu_gfm_axis} evaluates criteria-level transfer on TMU-GFM. SOME performs better across all axes on all pairs, which is expected given its in-domain axis-level supervision. However, when restricting the evaluation to confident pairs with human score differences of at least 1.0, SURE achieves reasonable zero-shot accuracy on grammaticality and fluency. Faithfulness remains the weakest axis, reflecting the difficulty of aligning our faithfulness criterion with TMU-GFM's meaning-preservation annotations.

\subsection{Additional Criteria-Pair Diagnostics} \label{app:criteria_scatter}
\begin{figure}[t!]
    \centering
    \includegraphics[width=\columnwidth, clip, trim=5 5 5 5]{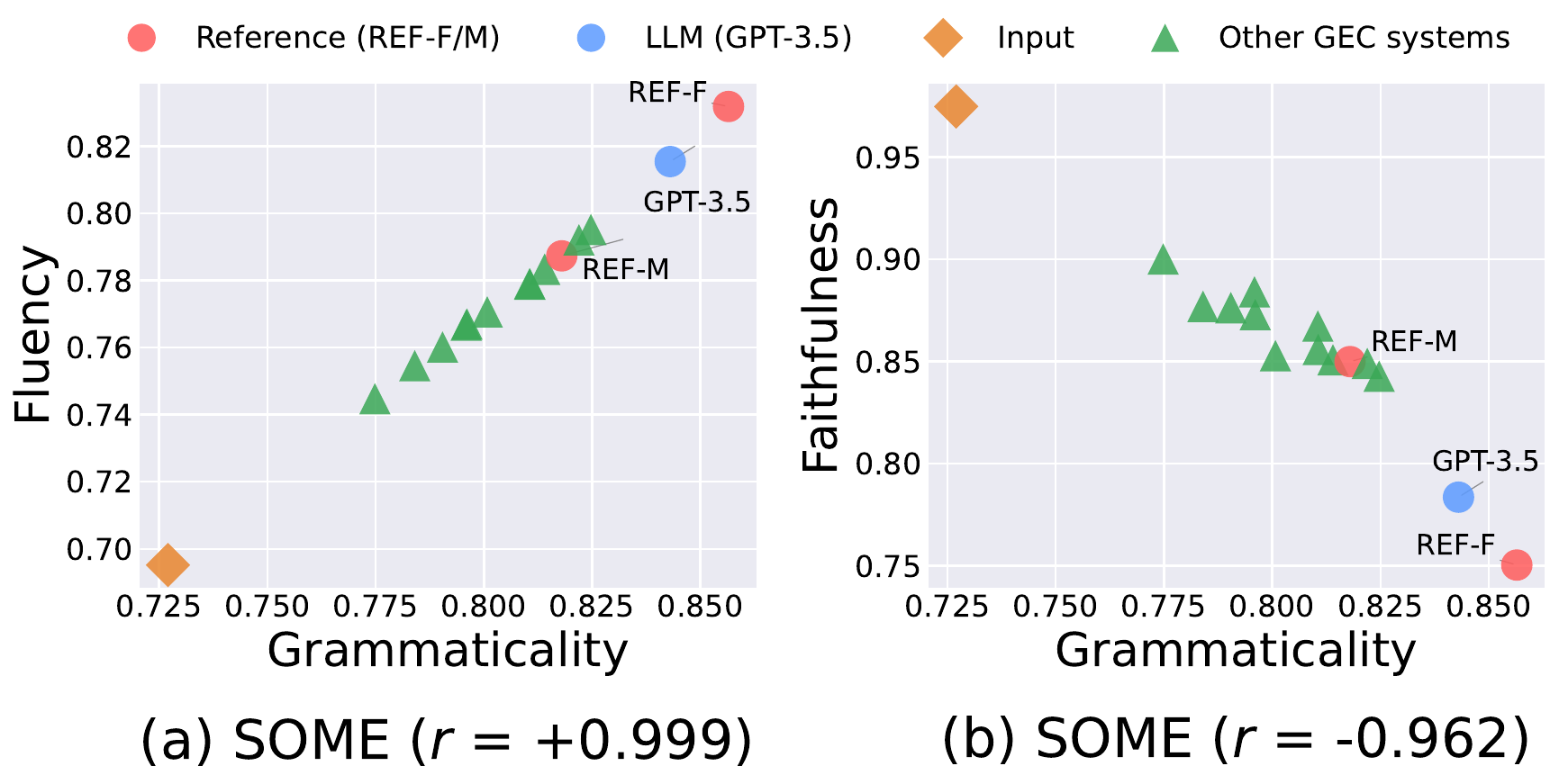}
    \caption{Additional criteria-pair correlations for SOME.}
\label{fig:criteria_SOME}
\end{figure}

\begin{figure}[t!]
    \centering
    \includegraphics[width=\columnwidth, clip, trim=5 5 5 5]{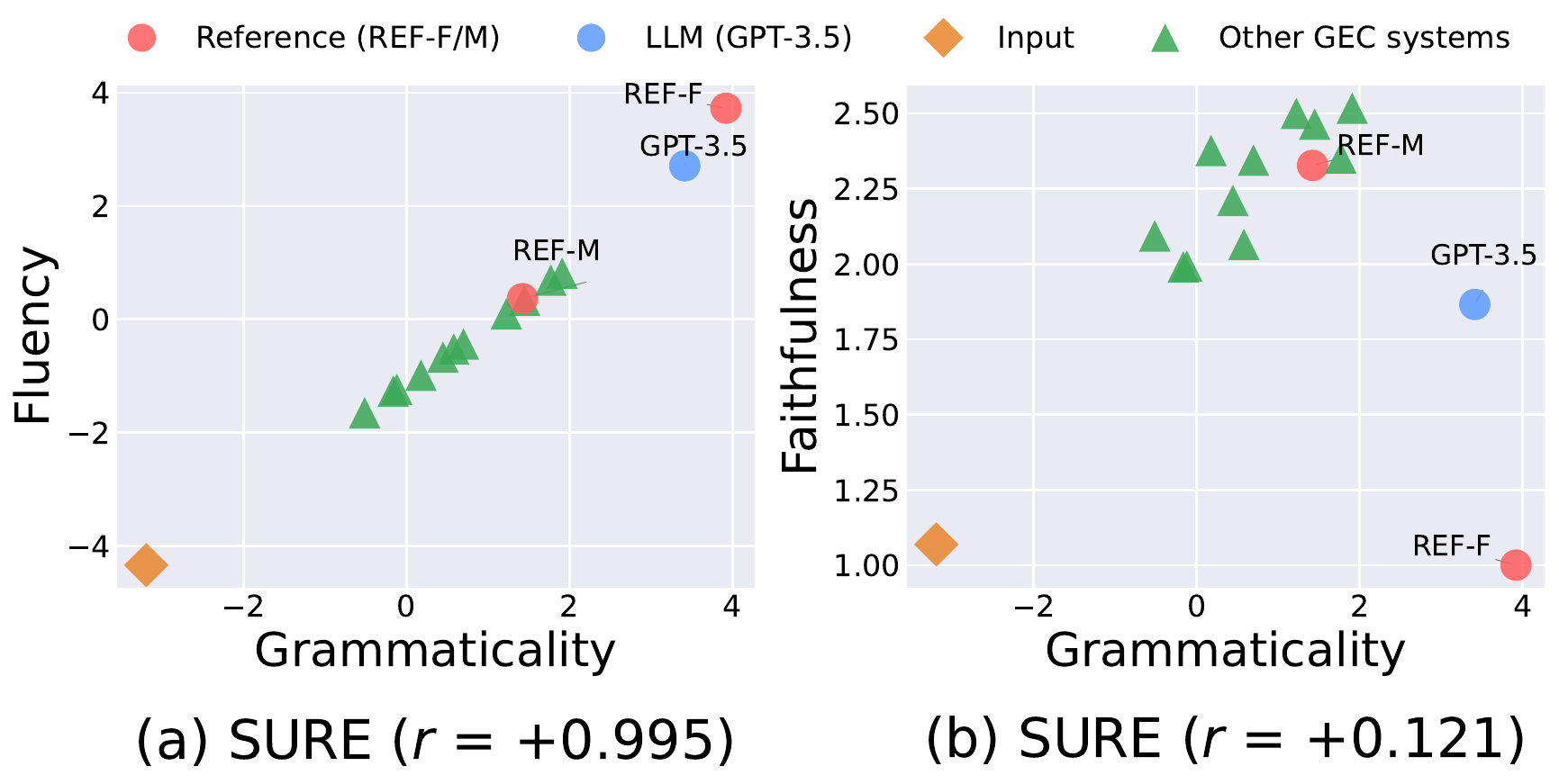}
    \caption{Additional criteria-pair correlations for SURE.}
\label{fig:criteria_SURE}
\end{figure}

Figures~\ref{fig:criteria_SOME} and~\ref{fig:criteria_SURE} report the remaining criteria-pair correlations, using SOME's meaning-preservation score as the counterpart of faithfulness. 
Both SOME and SURE show a near-perfect positive correlation between grammaticality and fluency, indicating that the two criteria capture closely related aspects of correction quality. The distinction lies in faithfulness-related pairs: SOME shows a strong negative correlation between grammaticality and faithfulness, mirroring the faithfulness--fluency entanglement observed in the main analysis. In contrast, SURE keeps grammaticality and faithfulness much less correlated, indicating that faithfulness is not simply determined by either grammaticality or fluency. These results suggest that SURE provides a more separated criteria space for diagnosing system-level correction behavior.

\raggedbottom
\section{Prompt Templates} \label{app:prompts}
\subsection{Correction Generation Prompts}
\begin{promptbox}{Prompt for Minimal-Edit Correction}
You are a careful English writing assistant.
\vspace{0.3em}

Correct the following sentence:
\vspace{0.3em}

Source: \\
\{source\}

\vspace{0.3em}
Return only the corrected sentence on one line, with no commentary.
\end{promptbox} % 보수적 (minimal edit 스타일, temperature=0.7)
\begin{promptbox}{Prompt for Rewrite-Oriented Correction}
You are a careful English writing assistant.

\vspace{0.3em}

Rewrite the following sentence so that it sounds natural and fluent in English.
You may restructure clauses, change word order, swap synonyms, or split/merge
clauses as long as the original meaning is preserved.
Do not add information that is not in the source.
Aim for a noticeably different surface form from a minimal-edit correction.
\vspace{0.3em}

Source:\\
\{source\}

\vspace{0.3em}
Return only the rewritten sentence on one line, with no commentary.
\end{promptbox} % 적극적 rewrite (style diversity, temperature=1.1)
\raggedbottom
\subsection{GPT-4.1 Evaluation Prompts}
\begin{promptbox}{GPT-4.1-E Evaluation Prompt}
You are evaluating English grammatical error corrections.
\vspace{0.5em}

Rank the \{n\} candidates by how well their \textbf{EDITS} (changes vs. the source) fix the source's errors.
\vspace{0.5em}

Focus only on the quality of the edits:
\begin{itemize}
    \item Reward edits that correctly fix grammatical, fluency, or usage errors.
    \item Penalize edits that introduce new errors, change the meaning, or make unnecessary changes.
    \item Do not reward a candidate simply for being fluent if its edits are not appropriate.
    \item If two candidates are similar, prefer the one that fixes more source errors with fewer unnecessary edits.
\end{itemize}

\vspace{0.5em}
Source: \\
\{source\}

\vspace{0.5em}
Candidates: \\
\{candidate\_block\}

\vspace{0.5em}
Output only the ranking from best to worst using candidate IDs.
\end{promptbox}
\begin{promptbox}{GPT-4.1-S Evaluation Prompt}
You are evaluating English sentences as grammatical corrections of a source.
\vspace{0.5em}

Rank the \{n\} candidates by \textbf{OVERALL sentence quality}.
\vspace{0.5em}

Focus on the final corrected sentence:
\begin{itemize}
    \item Reward candidates that are grammatical, fluent, natural, and clear.
    \item Penalize candidates that contain grammar errors, awkward wording, or unnatural phrasing.
    \item Penalize candidates that distort or omit important meaning from the source.
    \item If two candidates are similar, prefer the one that is more natural and faithful.
\end{itemize}

\vspace{0.5em}
Source: \\
\{source\}

\vspace{0.5em}
Candidates: \\
\{candidate\_block\}

\vspace{0.5em}
Output only the ranking from best to worst using candidate IDs.
\end{promptbox}

\begin{figure*}[t]
\centering
\subsection{Pairwise GEC Annotation Prompt}
\begin{promptbox}{Prompt for Pairwise GEC Annotation}
\small

You are an expert linguistic annotator for grammatical error correction (GEC).
You compare two candidate corrections of the same source sentence on multiple
axes. You always reply with a single JSON object that follows the requested
schema exactly. Do not include commentary outside the JSON.

\vspace{0.6em}

\textbf{Source sentence:}

\texttt{\{source\}}

\vspace{0.4em}
\textbf{Known errors in the source} extracted by ERRANT against a minimal-edit
reference; treat these as the ground-truth list of errors that should be addressed:

\texttt{\{error\_block\}}

\vspace{0.4em}
\textbf{Candidate A:}

\texttt{\{cand\_a\}}

\vspace{0.4em}
\textbf{Candidate B:}

\texttt{\{cand\_b\}}

\vspace{0.6em}
\textbf{Your task}

For each candidate, judge whether it \textsc{resolved} each listed error.
\begin{itemize}
    \item \texttt{"resolved"}: the error is corrected and the result is grammatical.
    \item \texttt{"missed"}: the error is still present or re-introduced.
    \item \texttt{"partial"}: attempted but incomplete, or introduces a new issue.
\end{itemize}

Then compare A vs B on three sentence-level axes:

\begin{itemize}
    \item \textbf{Grammaticality}: which candidate has fewer grammar errors w.r.t. the source.
    \item \textbf{Faithfulness}: which candidate better preserves the source's meaning and intent.
    \item \textbf{Fluency}: which candidate reads as more natural, idiomatic English.
\end{itemize}

Finally, give an \textsc{overall} preference for the better correction.

\vspace{0.6em}
\textbf{Important}

\begin{itemize}
    \item The amount of editing is not a criterion. Minimal edits and rewrites are equally valid;
    judge only on the criteria above.
    \item For each axis and overall, you must choose exactly one of \texttt{"A"} or \texttt{"B"}.
    Do not use \texttt{"tie"}. If the two candidates seem close on a criterion, still pick
    whichever is even marginally better.
\end{itemize}

\textbf{Reply with a single JSON object using this exact schema:}

\vspace{0.3em}
\begin{verbatim}
{
  "span_resolution": {
    "A": {"e1": "resolved|missed|partial", ...},
    "B": {"e1": "resolved|missed|partial", ...}
  },
  "axes": {
    "grammaticality": "A|B",
    "faithfulness": "A|B",
    "fluency": "A|B"
  },
  "overall": "A|B",
  "rationale": "<one short sentence>"
}
\end{verbatim}

\end{promptbox}

\label{fig:pairwise_gec_prompt}
\end{figure*}

\end{document}